\documentclass[11pt]{article}

\usepackage[final]{acl}

\usepackage{times}
\usepackage{latexsym}

\usepackage[T1]{fontenc}

\usepackage[utf8]{inputenc}

\usepackage{microtype}

\usepackage{inconsolata}

\usepackage{graphicx}

\usepackage{enumitem}
\usepackage{pifont}

\usepackage{dblfloatfix}
\usepackage{tikz}
\tikzset{forked edges/.style={}} 
\usepackage[edges]{forest}
\usepackage{xcolor}

\usepackage{longtable}
\usepackage{hyperref}
\usepackage{array}

\usepackage[most]{tcolorbox}
\usepackage{lipsum} 

\definecolor{selfevolagent_dark}{HTML}{37D2A6} 
\definecolor{selfevolagent_light}{HTML}{9BE9D3}
\definecolor{selfevolagent_lighter}{HTML}{CDF4E9}

\title{A Survey of Reinforcement Learning for Large Language Models under Data Scarcity: Challenges and Solutions}

\author{
\textbf{Zhiyin Yu}\textsuperscript{\rm $1,2$}, 
Yuchen Mou\textsuperscript{\rm $3$}, 
Juncheng Yan\textsuperscript{\rm $4$}, 
Junyu Luo\textsuperscript{\rm $1$}, 
Chunchun Chen\textsuperscript{\rm $5$},
Xing Wei\textsuperscript{\rm $5$},
\\
\textbf{Yunhui Liu}\textsuperscript{\rm $6$},
\textbf{Hongru Sun}\textsuperscript{\rm $7$},
\textbf{Yuxing Zhang}\textsuperscript{\rm $8$},
\textbf{Jun Xu}\textsuperscript{\rm $4$},
\textbf{Yatao Bian}\textsuperscript{\rm $3$},
\textbf{Ming Zhang}\textsuperscript{\rm $1$},
\textbf{Wei Ye}\textsuperscript{\rm $5$},\\
\textbf{Tieke He}\textsuperscript{\rm $6$},
\textbf{Jie Yang}\textsuperscript{\rm $7$},
\textbf{Guanjie Zheng}\textsuperscript{\rm $8$},
\textbf{Zhonghai Wu}\textsuperscript{\rm $1$ \textdagger},
\textbf{Bo Zhang}\textsuperscript{\rm $2$ \textdagger}, 
\textbf{Lei Bai}\textsuperscript{\rm $2$}, 
\textbf{Xiao Luo}\textsuperscript{\rm $9$}
\\
{\textsuperscript{\rm $1$}{Peking University}} 
{\textsuperscript{\rm $2$}Shanghai Artificial Intelligence Laboratory} 
{\textsuperscript{\rm $3$}National University of Singapore}\\
{\textsuperscript{\rm $4$}Nankai University}
{\textsuperscript{\rm $5$}Tongji University}
{\textsuperscript{\rm $6$}Nanjing University}
{\textsuperscript{\rm $7$}University of Wollongong}\\
{\textsuperscript{\rm $8$}Shanghai Jiao Tong University}
{\textsuperscript{\rm $9$}University of Wisconsin-Madison}  \\
\small\texttt{zhiyinyu25@stu.pku.edu.cn, zhangbo@pjlab.org.cn, wuzh@pku.edu.cn, xiao.luo@wisc.edu}\\
}

\makeatletter
\def\blfootnote{\xdef\@thefnmark{}\@footnotetext}
\makeatother

\begin{document}
\maketitle
\begin{abstract}
Reinforcement learning (RL) has emerged as a powerful post-training paradigm for enhancing the reasoning capabilities of large language models (LLMs). However, reinforcement learning for LLMs faces substantial data scarcity challenges, including the limited availability of high-quality external supervision and the constrained volume of model-generated experience. These limitations make data-efficient reinforcement learning a critical research direction. In this survey, we present the first systematic review of reinforcement learning for LLMs under data scarcity. We propose a bottom-up hierarchical framework built around three complementary perspectives: the data-centric perspective, the training-centric perspective, and the framework-centric perspective. We develop a taxonomy of existing methods, summarize representative approaches in each category, and analyze their strengths and limitations. Our taxonomy aims to provide a clear conceptual foundation for understanding the design space of data-efficient RL for LLMs and to guide researchers working in this emerging area. We hope this survey offers a comprehensive roadmap for future research and inspires new directions toward more efficient and scalable reinforcement learning post-training for LLMs.
\end{abstract}

\blfootnote{
\textsuperscript{\textdagger} Corresponding authors. 

{\url{https://github.com/YuZhiyin/Data-Efficient-RL}}\\
}

\section{Introduction}

Large language models (LLMs) have demonstrated remarkable capabilities across various domains, including mathematical reasoning~\citep{cui2025processreinforcementimplicitrewards,guan2025rstarmath}, algorithmic programming~\citep{li2024promptinglargelanguagemodels,guo2024deepseekcoderlargelanguagemodel} and scientific research~\citep{internagentteam2025internagentagentscientist}. As a prevailing and promising paradigm for post-training, reinforcement learning (RL) has shown strong potential to further enhance the reasoning abilities of LLMs. Recent advancements such as DeepSeek-R1~\citep{deepseekai2025deepseekr1incentivizingreasoningcapability} and OpenAI-o1~\citep{openai2024openaio1card} indicate that RL-based post-training can elicit emergent behaviors including self-reflection~\citep{zeng2025simplerlzoo}, enabling LLMs to complete complex tasks~\citep{kimiteam2025kimik15scalingreinforcement,hu2025openreasonerzeroopensourceapproach}.

\begin{figure}[t]
    \centering
    \includegraphics[width=\linewidth]{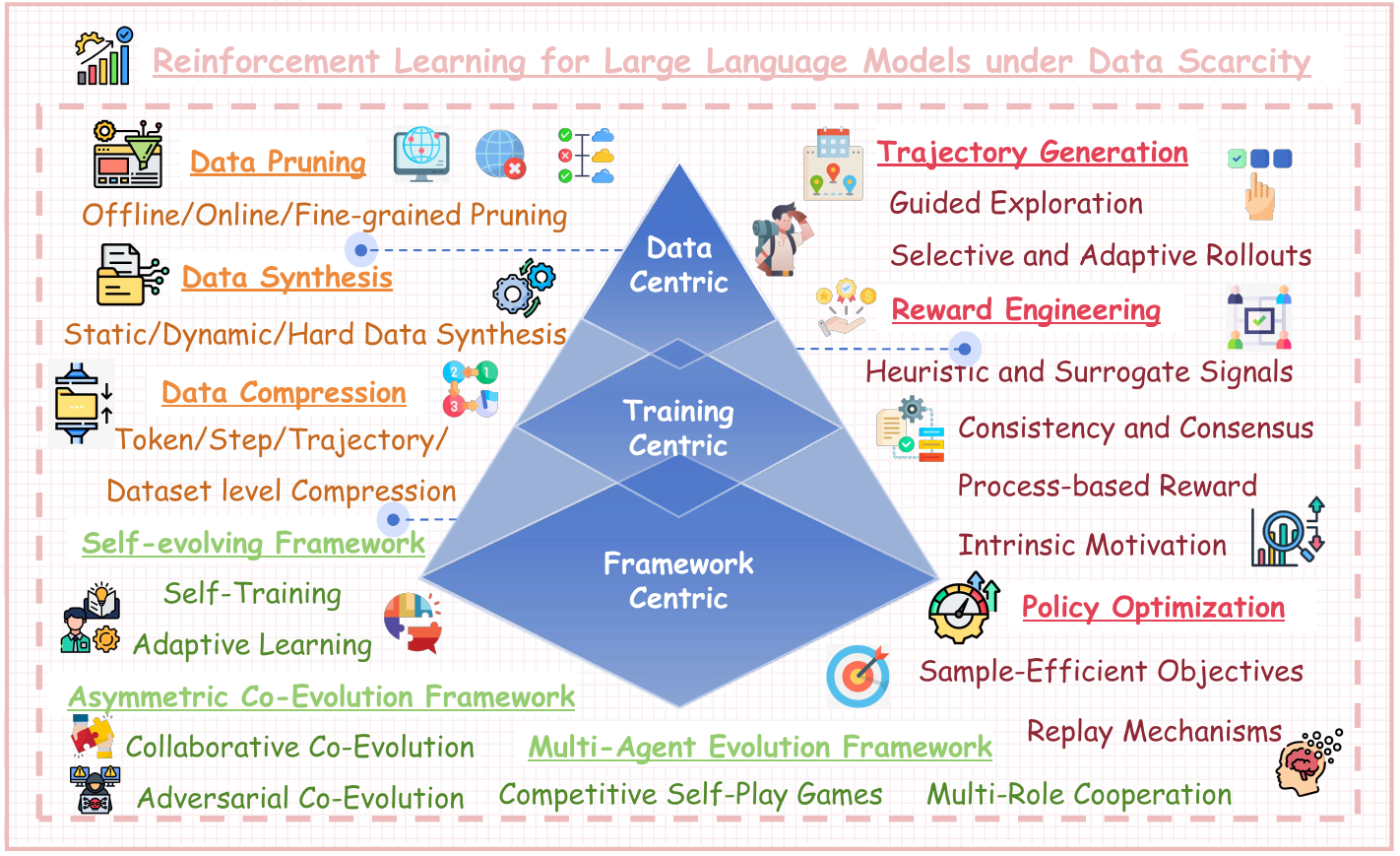}
    \caption{Overview of LLM-based reinforcement learning under data scarcity, illustrating data-, training-, and framework-centric perspectives.}
    \label{fig:overview}
    \vspace{-7mm}
\end{figure}

However, data scarcity has become a critical bottleneck constraining effective RL for LLMs. This challenge manifests in two complementary forms. Specifically, external data scarcity refers to the limited availability of high-cost supervised signals, such as fine-grained human feedback~\citep{wu2023finegrained}, preference data, expert annotations and step-by-step reasoning traces~\citep{xia-etal-2025-beyond}. By contrast, internal data scarcity arises from constraints on model-generated interactions, including the number of rollouts, trajectory lengths, and exploration budgets. As noted by \citet{jones2024ai}, “the AI revolution is running out of data”. Simply scaling training with more data or computational resources often yields diminishing returns and may still fail to produce strong performance~\citep{zuo2025ttrl}. Recently,~\citet{silver2025era} emphasized a shift toward "the era of experience", advocating a transition from heavy reliance on human supervision to approaches that allow models to evolve through experience. This perspective highlights the importance of studying reinforcement learning for LLMs under data scarcity. This paper provides the \textbf{first systematic survey} on RL for LLMs under data scarcity, offering a unified framework that organizes the fragmented research landscape. 


Existing research has explored various approaches to unlocking the potential of reinforcement learning for LLMs under data scarcity, including improving the utilization of limited human supervision~\citep{fang2025serl} and enhancing the efficiency of model-generated experience~\citep{wang2025beyond}. However, this field still lacks a systematic and unified survey. To bridge this gap, we review reinforcement learning under data scarcity from three complementary perspectives. Specifically, we introduce a conceptual framework consisting of data-centric, training-centric, and framework-centric perspectives (as illustrated in Figure~\ref{fig:overview}). Based on this framework, we design a taxonomy to categorize existing methods, summarize core techniques, and identify promising directions for future exploration. We hope this survey can serve as a roadmap for researchers and inspire continued progress in data-efficient RL for LLMs.

\noindent \textit{\textbf{Differences from Previous Surveys.}} Recently, several surveys have focused on related but orthogonal themes, including LLM and agentic RL~\citep{zhang2025surveyreinforcementlearninglarge,zhang2025landscapeagenticreinforcementlearning}, self-evolving agents~\citep{tao2024surveyselfevolutionlargelanguage,fang2025comprehensivesurveyselfevolvingai,gao2026surveyselfevolvingagentswhat}, and data-efficient post-training~\citep{luo-etal-2025-survey}. These surveys provide valuable insights but do not systematically study RL for LLMs under data scarcity. We address this gap with a bottom-up hierarchical framework and a unified analysis of existing methods through the lens of data scarcity.

\section{Taxonomy}
In this section, we present a taxonomy that categorizes reinforcement learning under data scarcity from three perspectives, as shown in Figure~\ref{fig:taxonomy}.

\begin{itemize}[left=0pt]
\vspace{-2mm}
\item \textit{\textbf{Level 1: Data-Centric Perspective:}} \textit{Optimizing the data itself to maximize usable information before, during, and after RL}. \ding{182} Data Pruning: Identifying the most informative subset from raw training data. 
\ding{183} Data Synthesis: Expanding or enriching the effective training distribution. 
\ding{184} Data Compression: Compressing tokens, reasoning steps, trajectories, or entire datasets while preserving essential information density.
\item \textit{\textbf{Level 2: Training-Centric Perspective:}} \textit{Improving how RL generates trajectories, evaluates rewards, and updates policies under scarce data}. 
\ding{182} Trajectory Generation: Improving exploration efficiency to reduce wasted rollouts. 
\ding{183} Reward Engineering: Enhancing the quality and granularity of feedback signals when annotations are limited. 
\ding{184} Policy Optimization: Increasing update efficiency and stability under constrained trajectory budgets.
\vspace{-2mm}
\item \textit{\textbf{Level 3: Framework-Centric Perspective:}} \textit{Designing evolving RL frameworks that reduce dependence on external data}. \ding{182} Self-Evolving Frameworks: Single-model systems that continually refine themselves through self-generated data and self-rewarding mechanisms. 
\ding{183} Asymmetric Co-Evolution Frameworks: Dual-agent architectures where agents optimize cooperative or adversarial objectives. 
\ding{184} Multi-Agent Evolution Frameworks: Systems of interacting agents that produce multi-objective training signals through cooperative or competitive dynamics.
\end{itemize}
These three perspectives collectively form a comprehensive and structured approach to RL under data scarcity, moving from curating the data, to improving the efficiency of trajectory usage, and ultimately to constructing autonomous frameworks capable of continual self-improvement.

\definecolor{softblue}{RGB}{220,230,242}
\definecolor{softgreen}{RGB}{226,239,218}
\definecolor{myyellow}{RGB}{255,242,204}
\definecolor{mypurple}{RGB}{229,224,236}
\definecolor{softred}{RGB}{242,220,219}
\definecolor{softgray}{RGB}{240,240,240}

\tikzstyle{leaf}=[draw=black, 
    rounded corners, minimum height=1em,
    text width=23em, 
    text opacity=1, align=left,
    fill opacity=.3, text=black, font=\scriptsize,
    inner xsep=5pt, inner ysep=3pt,
    ]
\tikzstyle{leaf1}=[draw=black, 
    rounded corners, minimum height=1em,
    text width=6em, 
    text opacity=1, align=center,
    fill opacity=.5, text=black, font=\scriptsize,
    inner xsep=3pt, inner ysep=3pt,
    ]
\tikzstyle{leaf2}=[draw=black, 
    rounded corners, minimum height=1em,
    text width=6em,  
    text opacity=1, align=center,
    fill opacity=.8, text=black, font=\scriptsize,
    inner xsep=3pt, inner ysep=3pt,
]

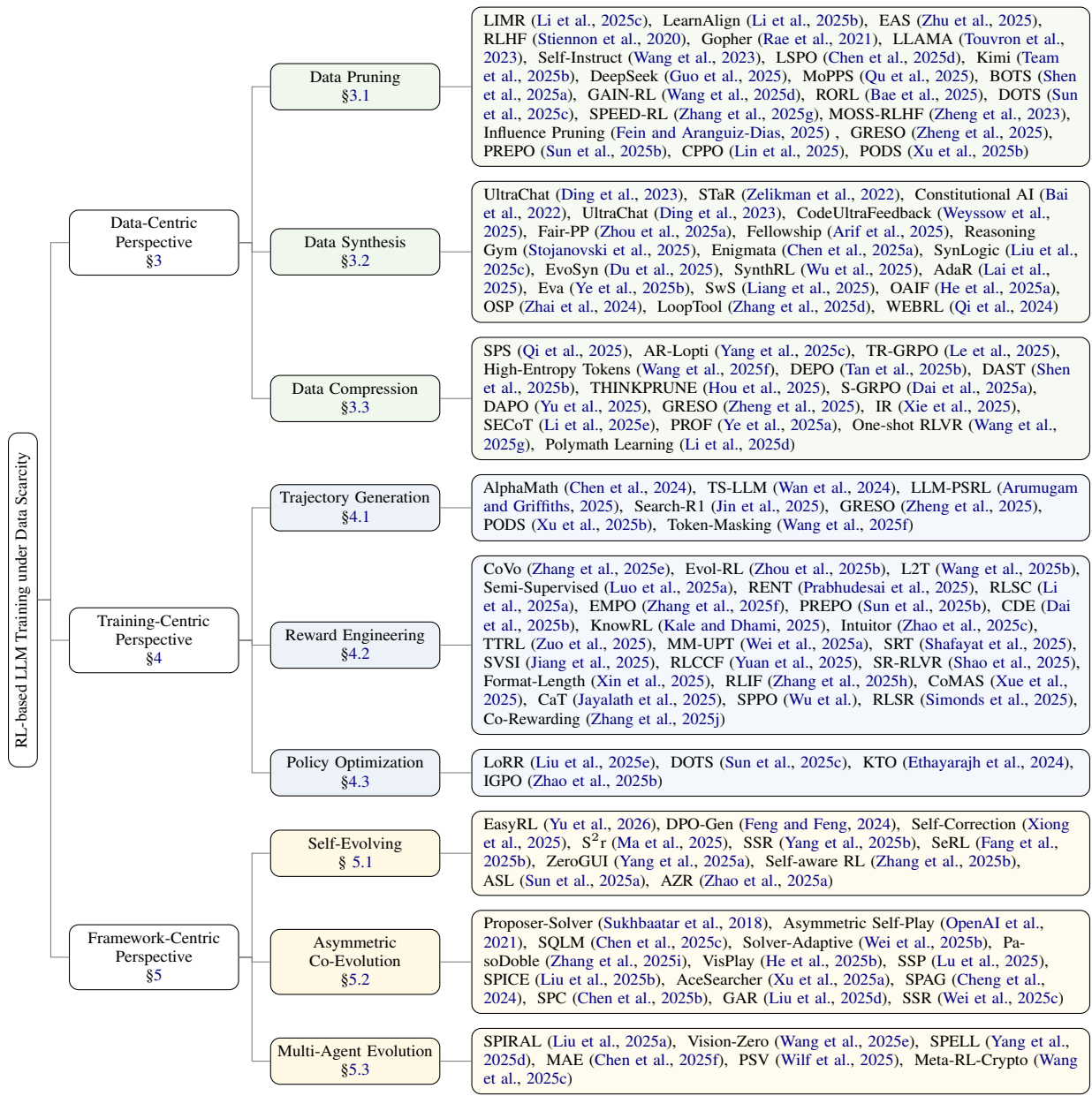
\begin{figure*}[t]
\centering
\begin{forest}
  for tree={
    forked edges,
    grow=east,
    reversed=true,
    anchor=center,
    parent anchor=east,
    child anchor=west,
    base=center,
    font=\scriptsize,
    rectangle,
    draw=black,
    edge=black!50, 
    rounded corners,
    minimum width=2em,
    s sep=6pt,
    inner xsep=3pt,
    inner ysep=2pt
  },
  where level=1{text width=7em}{},
  where level=2{text width=8em}{},
  [RL-based LLM Training under Data Scarcity, rotate=90, anchor=north, inner xsep=8pt, inner ysep=3pt, edge=black!50, draw=black
    [Data-Centric Perspective \\ \S \ref{sec:level1}, leaf2, fill=none
      [Data Pruning \\ \S \ref{subsec:L1DP}, leaf1, fill=softgreen
        [LIMR~\cite{LIMR}{, } 
        LearnAlign~\cite{LearnAlign}{, } 
        EAS~\cite{EAS}{, } 
        RLHF~\cite{RLHF}{, } 
        Gopher~\cite{Gopher}{, }
        LLAMA~\cite{LLAMA}{, } 
        Self-Instruct~\cite{Self-Instruct}{, } 
        LSPO~\cite{LSPO}{, } Kimi~\cite{kimiteam2025kimik15scalingreinforcement}{, } DeepSeek~\cite{deepseekai2025deepseekr1incentivizingreasoningcapability}{, } 
        MoPPS~\cite{MoPPS}{, } 
        BOTS~\cite{BOTS}{, } 
        GAIN-RL~\cite{GAIN-RL}{, } 
        RORL~\cite{RORL}{, } 
        DOTS~\cite{DOTS}{, } 
        SPEED-RL~\cite{SPEED-RL}{,}
        MOSS-RLHF~\cite{zheng2023secretsrlhflargelanguage}{,}
        Influence Pruning~\cite{Influence_Pruning} {, } 
        GRESO~\cite{GRESO}{, } 
        PREPO~\cite{PREPO}{, } 
        CPPO~\cite{CPPO}{, } 
        PODS~\cite{PODS}, leaf, fill=softgreen]
      ]
      [Data Synthesis \\ \S \ref{subsec:L1DS}, leaf1, fill=softgreen
        [UltraChat~\cite{UltraChat}{, } 
        STaR~\cite{STaR}{, } 
        Constitutional AI~\cite{Constitutional_AI}{, } 
        UltraChat~\cite{UltraChat}{, } 
        CodeUltraFeedback~\cite{CodeUltraFeedback}{, } 
        Fair-PP~\cite{FAIR_PP}{, } 
        Fellowship~\cite{Fellowship}{, } 
        Reasoning Gym~\cite{ReasoningGYM}{, } 
        Enigmata~\cite{Enigmata}{, } 
        SynLogic~\cite{SynLogic}{, } 
        EvoSyn~\cite{EvoSyn}{, }
        SynthRL~\cite{SynthRL}{, }
        AdaR~\cite{AdaR}{, }
        Eva~\cite{Eva}{, } 
        SwS~\cite{SwS}{, } 
        OAIF~\cite{OAIF}{, } 
        OSP~\cite{OSP}{, } 
        LoopTool~\cite{looptool}{, } 
        WEBRL~\cite{webrl}, leaf, fill=softgreen]
      ]
      [Data Compression \\ \S \ref{subsec:L1DC}, leaf1, fill=softgreen
        [SPS~\cite{shallow}{, } 
        AR-Lopti~\cite{arlopti}{, } 
        TR-GRPO~\cite{TR-GRPO}{, } 
        High-Entropy Tokens~\cite{wang2025beyond}{, } 
        DEPO~\cite{depo}{, } 
        DAST~\cite{dast}{, } 
        THINKPRUNE~\cite{thinkprune}{, } 
        S-GRPO~\cite{sgrpo}{, } 
        DAPO~\cite{dapo}{, } 
        GRESO~\cite{GRESO}{, } 
        IR~\cite{xie2025interleaved}{, } SECoT~\cite{li2025compressingchainofthoughtllmsstep}{, }
        PROF~\cite{prof}{, }  
        One-shot RLVR~\cite{oneshot}{, } 
        Polymath Learning~\cite{li2025one}, leaf, fill=softgreen]
      ]
    ]
    [Training-Centric Perspective \\ \S \ref{sec:level2}, leaf2, fill=none
      [Trajectory Generation \\ \S \ref{subsec:L2TG}, leaf1, fill=softblue
        [AlphaMath~\cite{chen2024alphamath}{, } 
        TS-LLM~\cite{wan2024alphazero}{, } 
        LLM-PSRL~\cite{arumugam2025toward}{, } 
        Search-R1~\cite{jin2025search}{, } 
        GRESO~\cite{GRESO}{, } 
        PODS~\cite{PODS}{, } 
        Token-Masking~\cite{wang2025beyond}, leaf, fill=softblue]
      ]
      [Reward Engineering \\ \S \ref{subsec:L2RE}, leaf1, fill=softblue
        [CoVo~\cite{zhang2025consistentpathsleadtruth}{, }
        Evol-RL~\cite{zhou2025evolvinglanguagemodelslabels}{, } L2T~\cite{wang2025learning}{, } 
        Semi-Supervised~\cite{luo2025semi}{, } 
        RENT~\cite{prabhudesai2025maximizing}{, } 
        RLSC~\cite{li2025confidence}{, } 
        EMPO~\cite{zhang2025right}{, } 
        PREPO~\cite{PREPO}{, } 
        CDE~\cite{dai2025cde}{, }
        KnowRL~\cite{kale2025knowrl}{, } 
        Intuitor~\cite{zhao2025learning}{, } 
        TTRL~\cite{zuo2025ttrl}{, } 
        MM-UPT~\cite{weifirst}{, } 
        SRT~\cite{shafayat2025largereasoningmodelsselftrain}{, } 
        SVSI~\cite{jiang2025semantic}{, } 
        RLCCF~\cite{yuan2025wisdom}{, } 
        SR-RLVR~\cite{shao2025spurious}{, } 
        Format-Length~\cite{xin2025surrogate}{, }
        RLIF~\cite{zhang2025no}{, }
        CoMAS~\cite{xue2025comas}{, } 
        CaT~\cite{jayalath2025compute}{, } 
        SPPO~\cite{wuself}{, } 
        RLSR~\cite{simonds2025self}{, } 
        Co-Rewarding~\cite{zhang2025co}, leaf, fill=softblue]
      ]
      [Policy Optimization \\ \S \ref{subsec:L2PO}, leaf1, fill=softblue
        [LoRR~\cite{liu2025sample}{, } 
        DOTS~\cite{DOTS}{, } 
        KTO~\cite{ethayarajh2024kto}{, } 
        IGPO~\cite{zhao2025inpainting}, leaf, fill=softblue]
      ]
    ]
    [Framework-Centric Perspective \\ \S \ref{sec:level3}, leaf2, fill=none
      [Self-Evolving \\ \S ~\ref{subsec:L3SEF}, leaf1, fill=myyellow
        [EasyRL~\cite{yu2026easy}{, }DPO-Gen~\cite{feng2024extremelydataefficientgenerativellmbased}{, } 
        Self-Correction~\cite{xiong2025selfrewardingcorrectionmathematicalreasoning}{, } 
        S$^2$r~\cite{ma2025s2rteachingllmsselfverify}{, } 
        SSR~\cite{yang2025ssrzerosimpleselfrewardingreinforcement}{, } 
        SeRL~\cite{fang2025serl}{, } 
        ZeroGUI~\cite{yang2025zeroguiautomatingonlinegui}{, } 
        Self-aware RL~\cite{zhang2025pathselfevolvinglargelanguage}{, } 
        ASL~\cite{sun2025agenticselflearningllmssearch}{, } AZR~\cite{zhao2025absolutezeroreinforcedselfplay}, leaf, fill=myyellow]
      ]
      [Asymmetric Co-Evolution \\ \S \ref{subsec:L3ACF}, leaf1, fill=myyellow
        [Proposer-Solver~\cite{sukhbaatar2018intrinsic}{, }
        Asymmetric Self-Play~\cite{openai2021asymmetricselfplayautomaticgoal}{, } 
        SQLM~\cite{chen2025selfquestioninglanguagemodels}{, } 
        Solver-Adaptive~\cite{wei2025learningposeproblemsreasoningdriven}{, } PasoDoble~\cite{zhang2025betterllmreasoningdualplay}{, } 
        VisPlay~\cite{he2025visplayselfevolvingvisionlanguagemodels}{, } 
        SSP~\cite{lu2025searchselfplaypushingfrontier}{, } 
        SPICE~\cite{liu2025spiceselfplaycorpusenvironments}{, } 
        AceSearcher~\cite{xu2025acesearcher}{, }
        SPAG~\cite{SPAG}{, }
        SPC~\cite{SPC}{, }
        GAR~\cite{GAR}{, }
        SSR~\cite{SSR}, leaf, fill=myyellow]
      ]
      [Multi-Agent Evolution \\ \S \ref{subsec:L3MAEF}, leaf1, fill=myyellow
        [SPIRAL~\cite{SPIRAL}{, } 
        Vision-Zero~\cite{Vision-Zero}{, } 
        SPELL~\cite{SPELL}{, } 
        MAE~\cite{MAE}{, } 
        PSV~\cite{PSV}{, } 
        Meta-RL-Crypto~\cite{Meta-RL-Crypto}, leaf, fill=myyellow]
      ]
    ]
  ]
\end{forest}
\caption{A taxonomy of RL-based LLM Training under Data Scarcity.}
\label{fig:taxonomy}
\vspace{-5mm}
\end{figure*}

\newcommand{\nosection}[1]{\vspace{3pt}\noindent\textbf{#1.}}

\section{Data-Centric Perspective}
\label{sec:level1}
\subsection{Data Pruning}
\label{subsec:L1DP}
Data pruning shrinks the effective training pool and accelerates training by selectively retaining informative samples. As shown in Figure~\ref{fig:level1}, we classify existing data pruning methods as: offline pruning, online pruning, and fine-grained pruning.

\noindent \textit{\textbf{Offline Pruning.}}
Heuristic pruning has been widely adopted across various LLM training paradigms~\cite{RLHF,Gopher,LLAMA,Self-Instruct}, where predefined rules are applied during data preprocessing from multiple perspectives,
making it a critical component of data curation~\cite{kimiteam2025kimik15scalingreinforcement,deepseekai2025deepseekr1incentivizingreasoningcapability}.
LIMR~\cite{LIMR} identifies high-impact prompts by measuring the alignment between each sample reward trajectory and the average learning curve. LearnAlign~\cite{LearnAlign} uses a learnability-weighted gradient-alignment score to select reasoning samples that are both learnable and representative. EAS~\cite{EAS} selects samples by integrating token-level predictive entropy along the generation trajectory.

\noindent \textit{\textbf{Online Pruning.}}
Online pruning dynamically selects high-information-density samples for rollout.
LSPO~\cite{LSPO} augments accuracy-based filtering with an additional length-based step. MMoPPS~\cite{MoPPS} estimates prompt difficulty as a latent success probability for adaptive prompt selection. BOTS~\cite{BOTS} further integrates explicit feedback from selected samples and implicit evidence from unselected samples to balance exploration and exploitation. GAIN-RL~\cite{GAIN-RL} predicts the magnitude of gradient updates using the angular concentration of pre-filling hidden states and schedules an easy-to-hard curriculum. RORL~\cite{RORL} estimates pass rates from online rollouts to identify prompts of moderate difficulty, thereby focusing updates on samples with higher pass rate variance. DOTS~\cite{DOTS} uses attention-based similarity to propagate difficulty scores from a reference subset to the full training pool. SPEED-RL~\cite{SPEED-RL} allocates training budget to moderately difficult samples by estimating empirical pass rates on a few samples. Beyond explicit sample selection,~\citet{zheng2023secretsrlhflargelanguage} investigates implicit data filtering strategies in PPO training, including reward clipping, response deduplication, and advantage normalization.


\vspace{-1mm}

\noindent \textit{\textbf{Fine-grained Pruning.}}
Fine-grained pruning reframes the problem toward precise pruning criteria and focuses selection on trajectories that are valuable to retain. Influence Pruning~\cite{Influence_Pruning} approximates influence functions using a conjugate gradient solver to estimate each sample’s effect on validation loss.
GRESO~\cite{GRESO} learns a reward-driven skipping policy to bypass prompts with zero reward variance, increasing the proportion of effective samples.
PREPO~\cite{PREPO} leverages prompt perplexity to adopt an easy-to-hard curriculum when the available number of prompts is limited. CPPO~\cite{CPPO} uses within-group advantages as a fine-grained learning signal to prune inefficient trajectories. PODS~\cite{PODS} performs down-sampling to obtain a subset of trajectories with the largest reward variance.

\subsection{Data Synthesis}
\label{subsec:L1DS}
\begin{figure}[t]
    \centering
    \includegraphics[width=\linewidth]{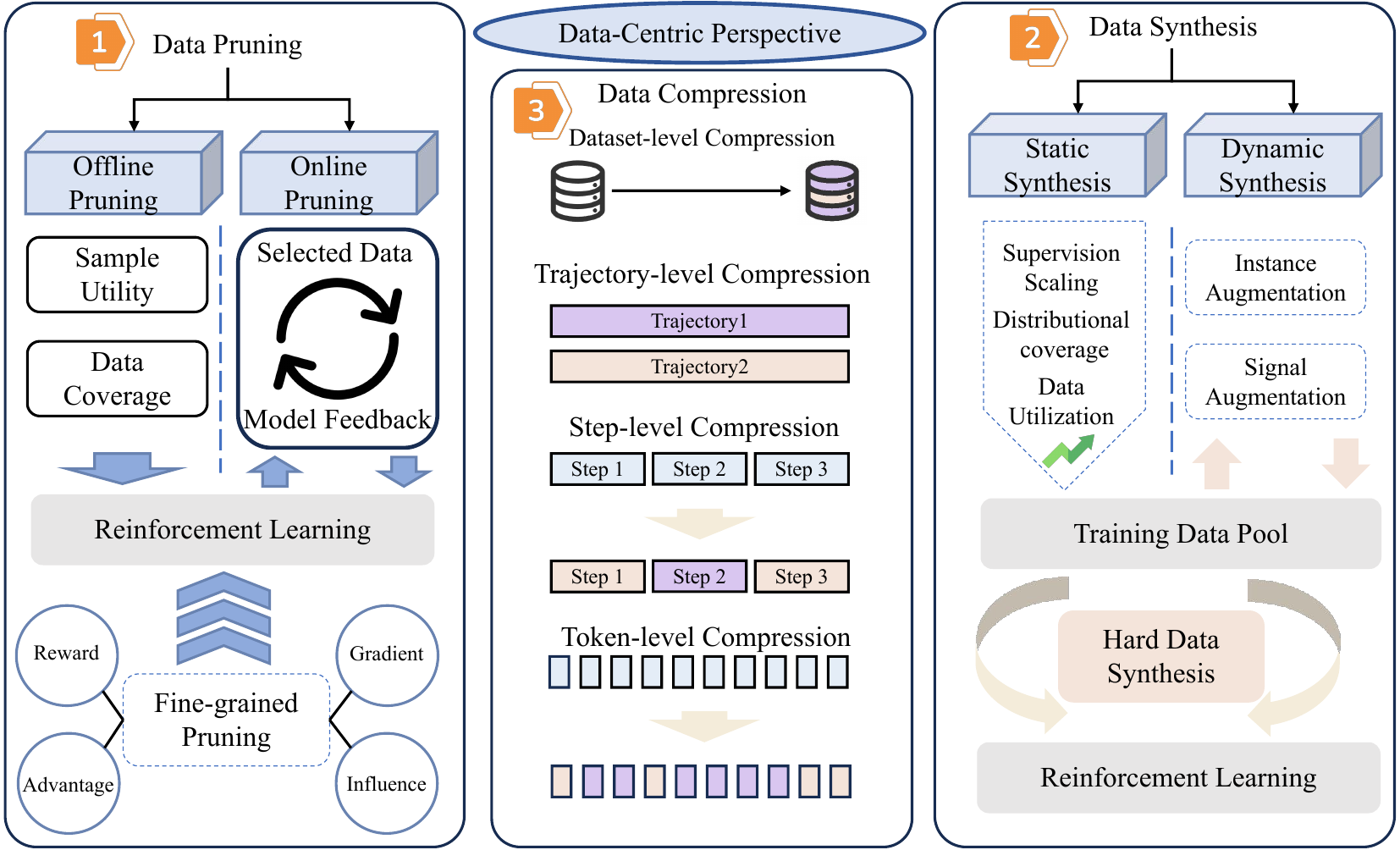}
    \caption{RL for LLMs in data-centric perspective.}

    \vspace{-5mm}
    \label{fig:level1}
\end{figure}

Data synthesis can expand the scale of supervision and improve distributional coverage under a constrained sample budget. As shown in Figure~\ref{fig:level1}, we classify existing methods as: static data synthesis before training, dynamic data synthesis during training, and hard data synthesis after training.

\noindent \textit{\textbf{Static Data Synthesis.}}
Static data synthesis~\cite{UltraChat,STaR} provides a solid data foundation~\cite{deepseekai2025deepseekr1incentivizingreasoningcapability} across different stages of LLM training. 
Constitutional AI~\cite{Constitutional_AI}, UltraFeedback~\cite{UltraFeedback}, and CodeUltraFeedback~\cite{CodeUltraFeedback} leverage strong LLMs to synthesize large scale preference and critique data with rich multi dimensional coverage. Meanwhile, Fair-PP~\cite{FAIR_PP} and Fellowship of the LLMs~\cite{Fellowship} further incorporate rule based constraints or multi agent collaboration. Reasoning Gym~\cite{ReasoningGYM}, Enigmata~\cite{Enigmata}, SynLogic~\cite{SynLogic}, and EvoSyn~\cite{EvoSyn} scale the synthesis of verifiable reasoning data within offline pipelines by designing diverse interaction schemes between generators and verifiers. Meanwhile, SynthRL~\cite{SynthRL} and AdaR~\cite{AdaR} focus on generating logically equivalent variant problems, with an emphasis on broader coverage of the training distribution and improved generalization.

\noindent \textit{\textbf{Dynamic Data Synthesis.}}
Dynamic data synthesis strengthens LLM training by continuously generating and augmenting data within the learning loop and can operate at different levels, such as instances and signals.
OSP~\cite{OSP} optimizes the policy using self-generated soft preference advantages, turning offline preference optimization into an online alignment pipeline while OAIF~\cite{OAIF} uses an online LLM annotator as a reward model to directly label on-policy response pairs.

\noindent \textit{\textbf{Hard Data Synthesis.}}
SwS~\cite{SwS} identifies the weakness problems that exhibit consistently low accuracy and degrading performance across epochs, and generates new training problems targeted at these weaknesses.
LoopTool~\cite{looptool} identifies and corrects label errors after each RL phase.
EVA~\cite{Eva} finds the most useful prompts based on reward signals and generates variants of these prompts for subsequent training.
WEBRL~\cite{webrl} automatically generates new tasks from a model's failed interactions and continuously incorporates these tasks into subsequent reinforcement learning stages.

\subsection{Data Compression}
\label{subsec:L1DC}
Data compression aims to minimize both data and computational costs in reinforcement learning. As illustrated in Figure~\ref{fig:level1}, existing data compression methods can be categorized into four levels: token, step, trajectory, and dataset.

\noindent \textit{\textbf{Token-level Compression.}}
Shallow Preference Signals~\cite{shallow} finds that many tokens are ineffective for reinforcement learning, indicating token-level compression is feasible.
AR-Lopti~\cite{arlopti} reweighting or isolating low-probability tokens during training. TR-GRPO~\cite{TR-GRPO} downweights low-probability tokens while placing greater emphasis on high-probability ones during gradient computation.
High-Entropy Minority Tokens~\cite{wang2025beyond} proposes updating the policy using only high-entropy tokens, which play a dominant role in determining reasoning branches and policy updates, while masking gradients from low-entropy tokens. 
DEPO~\cite{depo} separates reasoning tokens into efficient and inefficient segments, downweights the advantages of inefficient tokens, combined with difficulty-aware length penalties and advantage clipping for stable optimization.

\noindent \textit{\textbf{Step-level Compression.}}
DAST~\cite{dast} proposes a difficulty-aware reward mechanism based on a Token Length Budget, enabling models to automatically shorten reasoning for simple problems while preserving sufficient CoT for complex ones.
THINKPRUNE~\cite{thinkprune} encouraging models to actively prune redundant reasoning steps and progressively learn more efficient reasoning structures through iterative budget tightening.
S-GRPO~\cite{sgrpo} introduces a serial grouping strategy with decaying rewards, guiding models to recognize when intermediate reasoning is already sufficient and perform early exit along the reasoning trajectory.
Interleaved Reasoning~\cite{xie2025interleaved} trains models via reinforcement learning to alternate between generating intermediate answers and reasoning steps, thereby providing denser and more verifiable intermediate reward signals.
Step-Entropy-based CoT Compression~\cite{li2025compressingchainofthoughtllmsstep} shows that many low-entropy reasoning steps are redundant, and trains models to actively skip such steps during generation.

\begin{figure}[t]
    \centering
    \includegraphics[width=\linewidth]{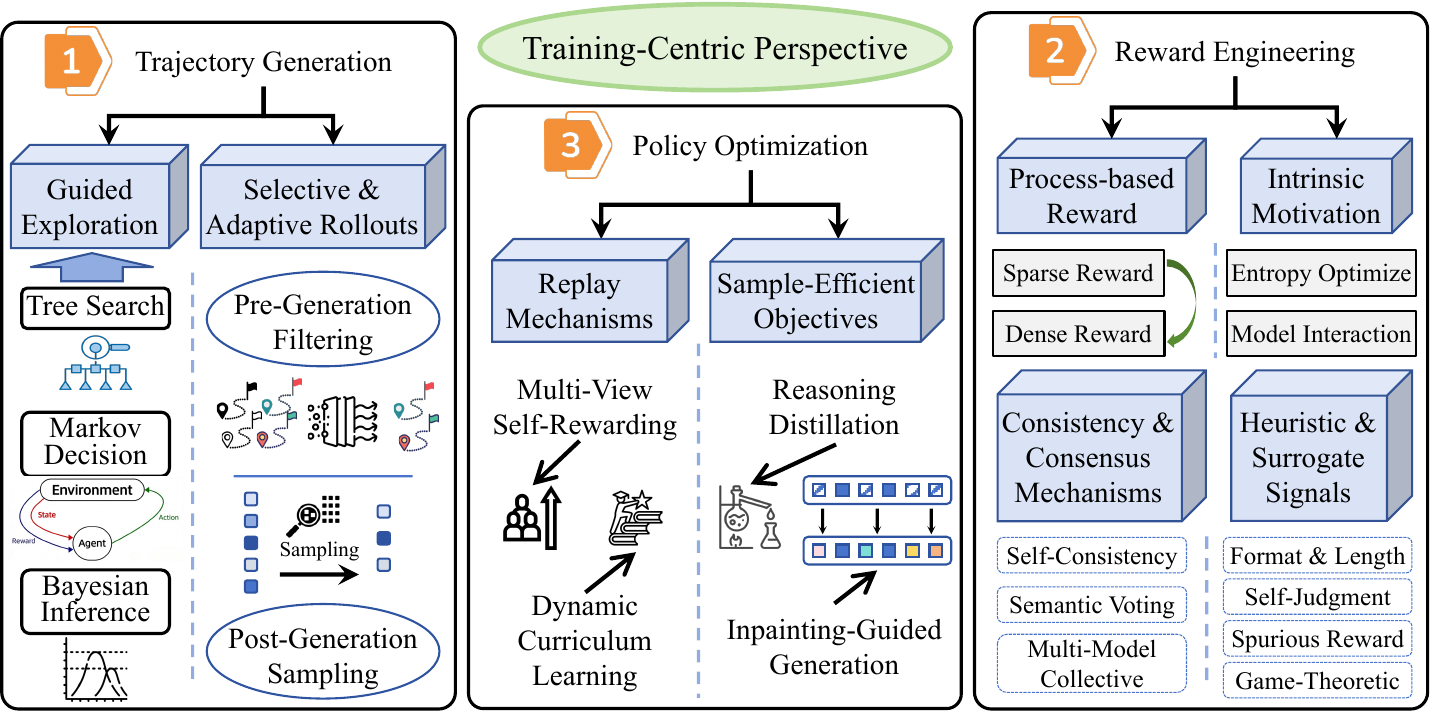}
    \caption{RL for LLMs in training-centric perspective.}

    \label{fig:training}
     \vspace{-5mm}
\end{figure}

\noindent \textit{\textbf{Trajectory-level Compression.}}
DAPO~\cite{dapo} oversamples prompts during training and filters out entire groups of trajectories that produce zero gradients, retaining only trajectories that yield non-zero gradient signals for optimization.
PROF~\cite{prof} performs trajectory-level ranking and filtering after rollout, retaining only those complete trajectories whose process rewards are consistent with outcome rewards.

\noindent \textit{\textbf{Dataset-level Compresion.}}
One-shot RLVR~\cite{oneshot} shows that the effective RLVR training set can be reduced to a single example while still achieving performance comparable to RLVR trained on datasets containing 7.5k samples. Polymath learning~\cite{li2025one} restricts the RL training set to a single example, selecting or synthesizing a high-information-density sample by explicitly covering salient reasoning skills.

\begin{tcolorbox}[
  colback=selfevolagent_light!20,
  colframe=selfevolagent_light!80,
  colbacktitle=selfevolagent_light!80,
  coltitle=black,
  title={\bfseries\fontfamily{ppl}\selectfont{Key takeaways}},
  boxrule=2pt,
  arc=5pt,
  drop shadow,
  parbox=false,
  before skip=5pt,
  after skip=5pt,
  left=5pt,   
  right=5pt,
]

\begin{itemize}[left=0pt,itemsep=0pt]
\item \textit{\textbf{Model-Dependent Valuation.}} Data value estimation depends on model capability, and overemphasizing high-contribution data may weaken long-tail learning.
\item \textit{\textbf{Adaptive Data Governance.}} Future work should jointly and adaptively schedule data pruning, compression, and synthesis.
\end{itemize}

\end{tcolorbox}


\section{Training-Centric Perspective}
\label{sec:level2}
From the perspective of training, researchers have addressed data dependence through three principal avenues: trajectory generation, reward engineering, and policy optimization (see Figure~\ref{fig:training}).

\subsection{Trajectory Generation}
\label{subsec:L2TG}
Policy updates rely on numerous trajectory samples. To avoid the unaffordable consumption of computational resources and time, researchers have shifted from unstructured random exploration toward guided exploration and selective rollouts.

\noindent \textit{\textbf{Guided Exploration.}}
Guided exploration seeks to improve sample efficiency through structured sampling. A major breakthrough in this area integrates planning algorithms such as Monte Carlo Tree Search (MCTS) into the LLM decoding process~\cite{chen2024alphamath,wan2024alphazero}, leveraging the structural advantages of search trees to increase the probability of generating valid samples. 
LLM-PSRL~\cite{arumugam2025toward} adopts a Bayesian posterior sampling approach through prompting engineering. It first samples a hypothesis from the posterior distribution, then takes optimal actions based on that hypothesis. 
A more advanced exploration direction uses external search tools as part of the exploration process. Search-R1~\cite{jin2025search} enables models to learn when to proactively initiate search queries in knowledge-blind regions. Such dynamic interleaving of search and reasoning trajectories encodes richer information-processing logic.

\noindent \textit{\textbf{Selective and Adaptive Rollouts.}}
Even with guided exploration, generating massive trajectories remains computationally expensive. Efficient rollout strategies~\cite{GRESO} have therefore become essential.
GRESO~\cite{GRESO} exemplifies \textit{pre-generation} filtering by maintaining historical states for each prompt and computing a skipping probability. Prompts predicted to yield low-information rollouts are directly bypassed. PODS~\cite{PODS}, a \textit{post-generation} sampling method, addresses the computational asymmetry between generation and update phases. Token-level masking methods~\cite{wang2025beyond} identify high-entropy tokens as the true forking points that determine reasoning path directions. By computing policy gradients only for high-entropy tokens, models not only maintain performance but can even achieve improvements.

\subsection{Reward Engineering}
\label{subsec:L2RE}
Evaluating trajectory quality is a central RL challenge. Sparse rewards complicate credit assignment, motivating automated process-based rewards and intrinsic motivation signals.

\noindent \textit{\textbf{Process-based Reward.}}
Process rewards aim to provide dense rewards for each reasoning step. In the absence of human-annotated process data, researchers leverage statistical properties of model outputs to synthesize process rewards.
CoVo~\cite{zhang2025consistentpathsleadtruth} observes that correct reasoning paths typically converge to the same answer, while incorrect paths diverge chaotically. It exploits consistency and volatility patterns to achieve a self-rewarding mechanism. Evol-RL~\cite{zhou2025evolvinglanguagemodelslabels} designs a novelty-promoting mechanism that rewards samples whose reasoning paths are semantically distant from existing paths yet still arrive at correct answers. L2T~\cite{wang2025learning} incorporates a compression penalty term that penalizes steps with low information gain but high token consumption. This reward design compels the model to learn efficient thinking, achieving maximum confidence improvement with minimal reasoning steps.

\noindent \textit{\textbf{Intrinsic Motivation.}}
When external feedback is unavailable, agents must rely on intrinsic motivation. Entropy-based metrics~\cite{luo2025semi} have become central to measuring intrinsic motivation. However, the community holds divergent views on how to leverage entropy. \textit{Entropy minimization} perspective~\cite{agarwal2025unreasonable} argues that pretrained models already contain most required knowledge, and errors primarily stem from decoding uncertainty. Therefore, RL should minimize entropy to make models more confident~\cite{prabhudesai2025maximizing,li2025confidence,zhang2025right}. 
However, \textit{Entropy maximization} perspective counters that solely minimizing entropy leads to premature convergence and overfitting. To push beyond capability boundaries, relative entropy must be maximized. PREPO~\cite{PREPO} and CDE~\cite{dai2025cde} reward rollouts with higher relative entropy (greater divergence from the old policy) to encourage exploration of new strategies. 
Beyond entropy, inter-model interactions also serve as important intrinsic rewards. SCoRe~\cite{kumartraining} and ReviewRL~\cite{zeng2025reviewrl} train models through multi-round and multi-agent RL to provide more reliable rewards.

\noindent \textit{\textbf{Consistency and Consensus Mechanisms.}}
In addition to entropy-driven exploration and multi-agent interaction, another powerful source of internal signals emerges from the self-consistency of the model's own response. 
For instance,
KnowRL~\cite{kale2025knowrl} incorporates introspection and consensus-based rewarding mechanisms to allow models to self-improve their boundary awareness with the internally-generated data. 
Intuitor~\cite{zhao2025learning} utilizes the confidence of the model itself as the sole reward signal, realizing fully unsupervised learning.
Majority voting reward methods~\cite{zuo2025ttrl,weifirst,shafayat2025largereasoningmodelsselftrain} estimate reward signals through majority voting, allowing models to self-evolve during test-time inference. 
SVSI~\cite{jiang2025semantic} leverages lightweight sentence embeddings to measure semantic similarity among generated responses, thereby constructing preference pairs for model training.
RLCCF~\cite{yuan2025wisdom} also uses self-consistency to weight the votes from multiple heterogeneous LLMs.

\noindent \textit{\textbf{Heuristic and Surrogate Signals.}}
Since consensus-based rewards require multiple deployments for each query, researchers turn to lighter heuristic rewards from individual responses.
\cite{shao2025spurious} uncovers that RL can improve mathematical reasoning in specific models even with spurious rewards. It shows that RL training activates reasoning patterns already learned during pretraining.
\cite{xin2025surrogate} suggests using format correctness and response length as surrogate reward signals.
Researchers also find that the efficacy of internal feedback-based RL depends on the model’s initial policy entropy~\cite{zhang2025no}.
CoMAS~\cite{xue2025comas} provide a co-evolution framework for multi-agent systems using intrinsic rewards, while \cite{zhang2025co} proposes a co-rewarding framework to provide complementary supervision on the data side and model side.
CaT~\cite{jayalath2025compute} synthesizes a reference answer from the model’s parallel inference-time rollouts and converts it into reference-free rewards.
SPPO~\cite{wuself} frames LLM alignment as a two-player constant-sum game and iteratively updates policies to approximate the Nash equilibrium.


\begin{figure}[t]
    \centering
    \includegraphics[width=\linewidth]{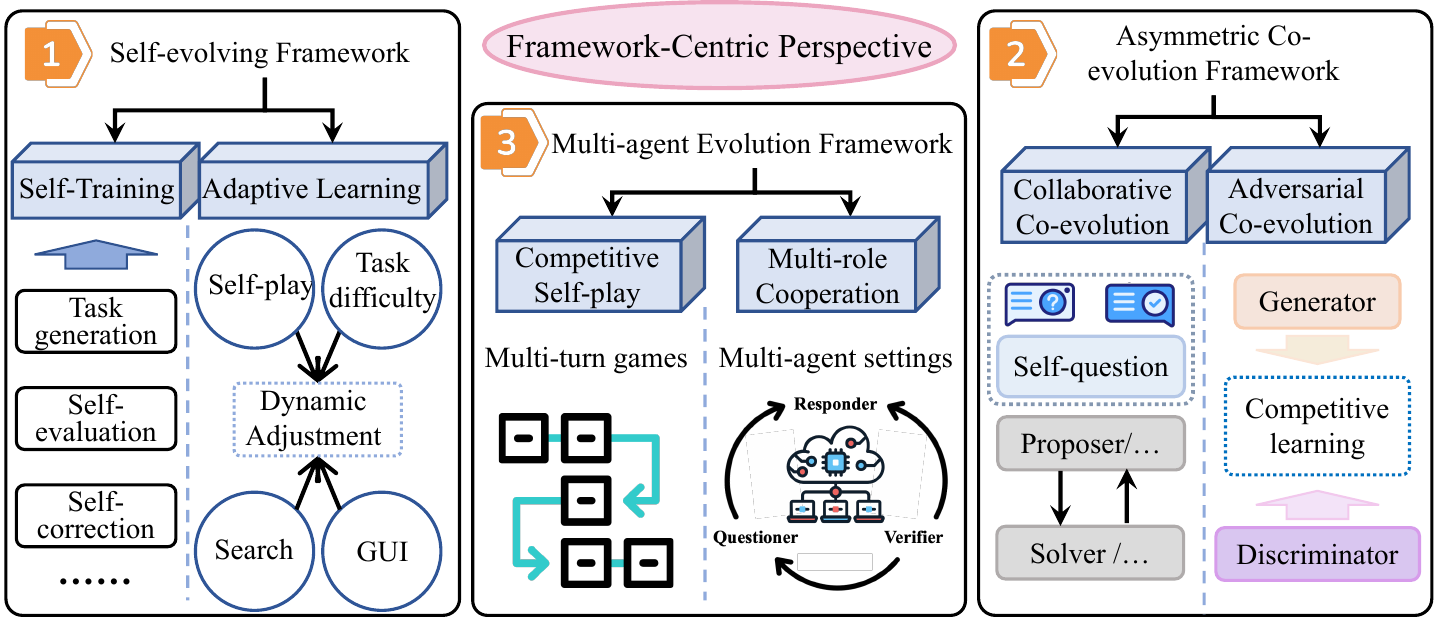}
    \caption{RL for LLMs in framework perspective.}

    \label{fig:framework}
    \vspace{-4mm}
\end{figure}

\subsection{Policy Optimization}
\label{subsec:L2PO}
The effectiveness of policy optimization determines whether the model can efficiently acquire reliable reasoning capabilities from generated trajectories and engineered rewards. 

\noindent \textit{\textbf{Replay Mechanisms.}}
To stabilize training and enhance sample efficiency, some methods revisit experience replay to adapt it to the unique structure of LLM reasoning trajectories.
\cite{liu2025sample} designs a multi-view self-rewarding mechanism to construct cross-validated intrinsic reward signals across diverse reasoning paths or problem formulations.
\cite{DOTS} presents a curriculum learning strategy with dynamic difficulty adjustment in reinforcement learning, enabling models to progressively advance from simple to complex problems in mathematical reasoning.

\noindent \textit{\textbf{Sample-Efficient Objectives.}}
Another pipeline focuses on redesigning the learning objective to maximize knowledge extraction from self-generated data.
KTO~\cite{ethayarajh2024kto} develops a distillation mechanism that leverages self-generated intermediate reasoning steps as supervision signals during training to enhance LLMs' reasoning abilities.
IGPO~\cite{zhao2025inpainting} introduces an inpainting-like controllable generation mechanism into diffusion language models to enable guidance and alignment of generated content.

\vspace{0.6em}
\begin{tcolorbox}[
  colback=selfevolagent_light!20,
  colframe=selfevolagent_light!80,
  colbacktitle=selfevolagent_light!80,
  coltitle=black,
  title={\bfseries\fontfamily{ppl}\selectfont{Key takeaways}},
  boxrule=2pt,
  arc=5pt,
  drop shadow,
  parbox=false,
  before skip=5pt,
  after skip=5pt,
  left=5pt,   
  right=5pt,
]

\begin{itemize}[left=0pt,itemsep=0pt]
\item \textit{\textbf{Mining Intrinsic Experience.}} The paradigm shifts from external supervision to mining intrinsic experiential value via guided exploration and endogenous signals.
\vspace{-2mm}
\item \textit{\textbf{Efficiency versus Noise.}} A core challenge is trading off sampling cost and signal-to-noise ratio of self-generated rewards, while preventing reward-hacking collapse.
\end{itemize}
\end{tcolorbox}

\section{Framework-Centric Perspective}
\label{sec:level3}
From a framework-centric perspective, RL under data scarcity leverages self-feedback to enable continuous evolution. We categorize methods into three paradigms: self-evolving, asymmetric co-evolution, and multi-agent evolution (see Figure~\ref{fig:framework}). 

\subsection{Self-evolving Framework}
\label{subsec:L3SEF}
Self-evolving utilizes a single model acting as both generator and evaluator, effectively closing the learning loop without external supervision.

\noindent \textit{\textbf{Self-Training.}}
Self-training methods evolve through iterative task generation and self-evaluation.~\citet{yu2026easy} enables self-evolving by progressively leveraging easy labeled data and harder unlabeled data via pseudo-labeling and RL. \citet{feng2024extremelydataefficientgenerativellmbased} combines Direct Preference Optimization with self-generated trajectories to reduce reliance on human labels. 
Similarly, self-correction frameworks~\cite{xiong2025selfrewardingcorrectionmathematicalreasoning} and self-verification mechanisms like \textsc{S$^2$r}~\cite{ma2025s2rteachingllmsselfverify} reinforce learning at both outcome and process levels. 
Recent works further refine this via self-judging rewards for machine translation~\cite{yang2025ssrzerosimpleselfrewardingreinforcement}, adaptive reward interpolation in RLER~\cite{tan2025diagnosingmitigatingbiasselfrewarding}, and bootstrapping via self-instruction in SeRL~\cite{fang2025serl}.


\noindent \textit{\textbf{Adaptive Learning.}}
These methods focus on dynamic strategy adjustment w.r.t environmental changes. 
ZeroGUI~\cite{yang2025zeroguiautomatingonlinegui} automates task generation for GUI agents to eliminate hand-crafted evaluations. 
Self-aware RL~\cite{zhang2025pathselfevolvinglargelanguage} enables capability prediction to proactively request data. ASL~\cite{sun2025agenticselflearningllmssearch} unifies task generation and execution in search environments. 
Similarly, AZR~\cite{zhao2025absolutezeroreinforcedselfplay} uses self-play to autonomously generate and solve training tasks.

\subsection{Asymmetric Co-Evolution Framework}
\label{subsec:L3ACF}
Asymmetric co-evolution typically involves two distinct agents to enhance learning through cooperative or adversarial interactions.

\noindent \textit{\textbf{Collaborative Co-Evolution.}}
This approach often pairs a proposer with a solver~\cite{sukhbaatar2018intrinsic,openai2021asymmetricselfplayautomaticgoal}, which utilizes intrinsic motivation for automatic curriculum generation. 
This dynamic extends to self-questioning~\cite{chen2025selfquestioninglanguagemodels}, difficulty adjustment~\cite{wei2025learningposeproblemsreasoningdriven}, and visual reasoning~\cite{he2025visplayselfevolvingvisionlanguagemodels}. 
Recent methods push the capability frontier by generating challenging queries via a Challenger~\cite{liu2025spiceselfplaycorpusenvironments} or unifying roles for search~\cite{lu2025searchselfplaypushingfrontier,xu2025acesearcher}. PasoDoble~\cite{zhang2025betterllmreasoningdualplay} further stabilizes training by decoupling updates between the proposer and solver.

\noindent \textit{\textbf{Adversarial Co-Evolution.}}
Adversarial frameworks often pit a generator against a discriminator. 
SPAG~\cite{SPAG} utilizes games like Adversarial Taboo to enforce information-reserved constraints. 
Verification capabilities are refined in SPC~\cite{SPC} via a "sneaky generator" designed to fool a critic, and in GAR~\cite{GAR} through a discriminator providing dense logical rewards. SSR~\cite{SSR} employs an injection-repair loop to construct a code repair curriculum via bug generation.


\subsection{Multi-Agent Evolution Framework}
\label{subsec:L3MAEF}
This paradigm generalizes self-play beyond binary interactions, utilizing complex game dynamics or specialized roles to internalize evaluation.

\noindent \textit{\textbf{Competitive Self-Play Games.}}
Competitive self-play allows LLMs to autonomously acquire reasoning skills by competing against themselves in structured, zero-sum environments. 
SPIRAL~\cite{SPIRAL} uses multi-turn games (e.g., poker) to incentivize systematic reasoning. 
Similarly, Vision-Zero~\cite{Vision-Zero} adapts competitive logic to visual domains via "Who is the Spy" games, achieving SOTA results in visual reasoning.

\noindent \textit{\textbf{Multi-Role Cooperation.}}
These models employ specialized roles (e.g., Proposer, Solver, Verifier) to address complex tasks. 
SPELL~\cite{SPELL} and MAE~\cite{MAE} utilize triplet structures to handle long-context and general reasoning. 
Others leverage formal verification signals~\cite{PSV} or hierarchical actor-judge architectures~\cite{Meta-RL-Crypto} to refine both policies and evaluation criteria within a closed loop.

\vspace{0.6em}
\begin{tcolorbox}[
  colback=selfevolagent_light!20,
  colframe=selfevolagent_light!80,
  colbacktitle=selfevolagent_light!80,
  coltitle=black,
  title={\bfseries\fontfamily{ppl}\selectfont{Key takeaways}},
  boxrule=2pt,
  arc=5pt,
  drop shadow,
  parbox=false,
  before skip=5pt,
  after skip=5pt,
  left=5pt,   
  right=5pt,
]

\begin{itemize}[left=0pt, itemsep=0pt]
  \item \textit{\textbf{Trade-offs.}} Self-evolving frameworks prioritize efficiency, while multi-agent methods trade cost for deeper reasoning.
  \vspace{-2mm}
    \item \textit{\textbf{De-biasing.}} Interactions in co-evolution introduce external signals, breaking "echo chambers" and reducing self-delusion.
    \vspace{-2mm}
    \item \textit{\textbf{Autogenous Curricula.}} These frameworks replace static datasets with dynamic generation, where task difficulty adaptively matches the model's evolving capability.
\end{itemize}

\end{tcolorbox}

\section{Challenges and Future Directions}
\(\triangleright\) \textit{\textbf{Reliability of Internal Rewards.}} LLM RL under data scarcity relies heavily on internal rewards~\citep{zuo2025ttrl} such as consistency, entropy and heuristic signals. However, these signals are often noisy and susceptible to reward hacking~\citep{shafayat2025largereasoningmodelsselftrain} or model collapse~\citep{shumailov2024ai}, making reliable credit assignments challenging. Future research should explore robust process-based signals~\citep{zhang2025consistentpathsleadtruth}, and hybrid reward designs~\citep{zhou2025evolvinglanguagemodelslabels} that remain stable even under scarce or noisy feedback.

\noindent \(\triangleright\) \textit{\textbf{Generalization to Unverifiable and Open-Ended Tasks.}} Most existing RL methods for LLMs under data scarcity focus on verifiable domains such as mathematics or coding~\citep{liu2025spiceselfplaycorpusenvironments,wei2025learningposeproblemsreasoningdriven} and are highly domain-dependent~\citep{zhang2025betterllmreasoningdualplay}. Future work should address unverifiable or subjective tasks (e.g., creative writing, open-ended dialogue~\citep{huang2025rzeroselfevolvingreasoningllm}, scientific discovery), real-world solving (e.g., world modeling and embodied agents~\citep{zhao2025absolutezeroreinforcedselfplay}), and generalization to out-of-distribution tasks.

\noindent \(\triangleright\) \textit{\textbf{Safety Risks in Self-Play Frameworks.}} Although self-play frameworks can generate data that reduce reliance on human-crafted tasks, they also introduce significant safety risks~\citep{wang-etal-2024-boosting-llm,wang-etal-2025-model}. For example, Llama-3.1-8B may exhibit “uh-oh moments” in chain-of-thought reasoning~\citep{zhao2025absolutezeroreinforcedselfplay}, and unsupervised self-evolution can propagate or even amplify biases from initial seed data~\citep{fang2025serl}. Under data-scarce conditions, models may further develop spurious generalization patterns. To mitigate these risks, future work should incorporate online filtering~\citep{chen2025selfquestioninglanguagemodels} and safety-aware training mechanisms to ensure reliable self-play processes.

\section{Conclusion}
We present a review of reinforcement learning for LLMs under data scarcity and introduce a bottom-up hierarchical taxonomy that categorizes existing approaches from data, training, and framework perspectives. Our survey reveals that addressing data scarcity goes beyond scaling supervised signals or interactive experience, but instead requires external data processing, effective internal data utilization during RL training, and evolving frameworks. We hope this survey provides a solid foundation for future research on data-efficient RL for LLMs.

\section{Limitations}
While this survey provides a first unified framework for reinforcement learning under data scarcity, it has limitations. Due to the rapid development of the field, the proposed framework requires timely updates to comprehensively cover emerging methods. We hope that this survey serves as inspiration for both theoretical and practical advancements in data-efficient reinforcement learning.

\section{Ethical Statement}
We follow the ACL Code of Ethics and maintain high standards of research integrity throughout this survey. As a literature review, our work does not involve human subjects, human annotation, or the collection of private or sensitive data. All referenced datasets, benchmarks, and models discussed in this survey are drawn from publicly available sources, and we cite them in accordance with their original usage and licensing conditions.

We aim to present a balanced and transparent analysis of existing methods, carefully summarizing their contributions, limitations, and potential risks. No conflicts of interest or sponsorship biases have been identified. We remain committed to addressing any ethical considerations raised during the review process and to promoting responsible research on data-efficient reinforcement learning for large language models.

\section{Acknowledgments}
The work of Zhiyin Yu, Bo Zhang, and Lei Bai was supported by the Shanghai Artificial Intelligence Laboratory. The authors thank the anonymous reviewers for their valuable comments and suggestions.

\bibliography{custom}

\clearpage
\appendix

\section{Statistics}
\label{app:statistics}
\begin{figure}[ht]
    \centering
    \includegraphics[width=\linewidth]{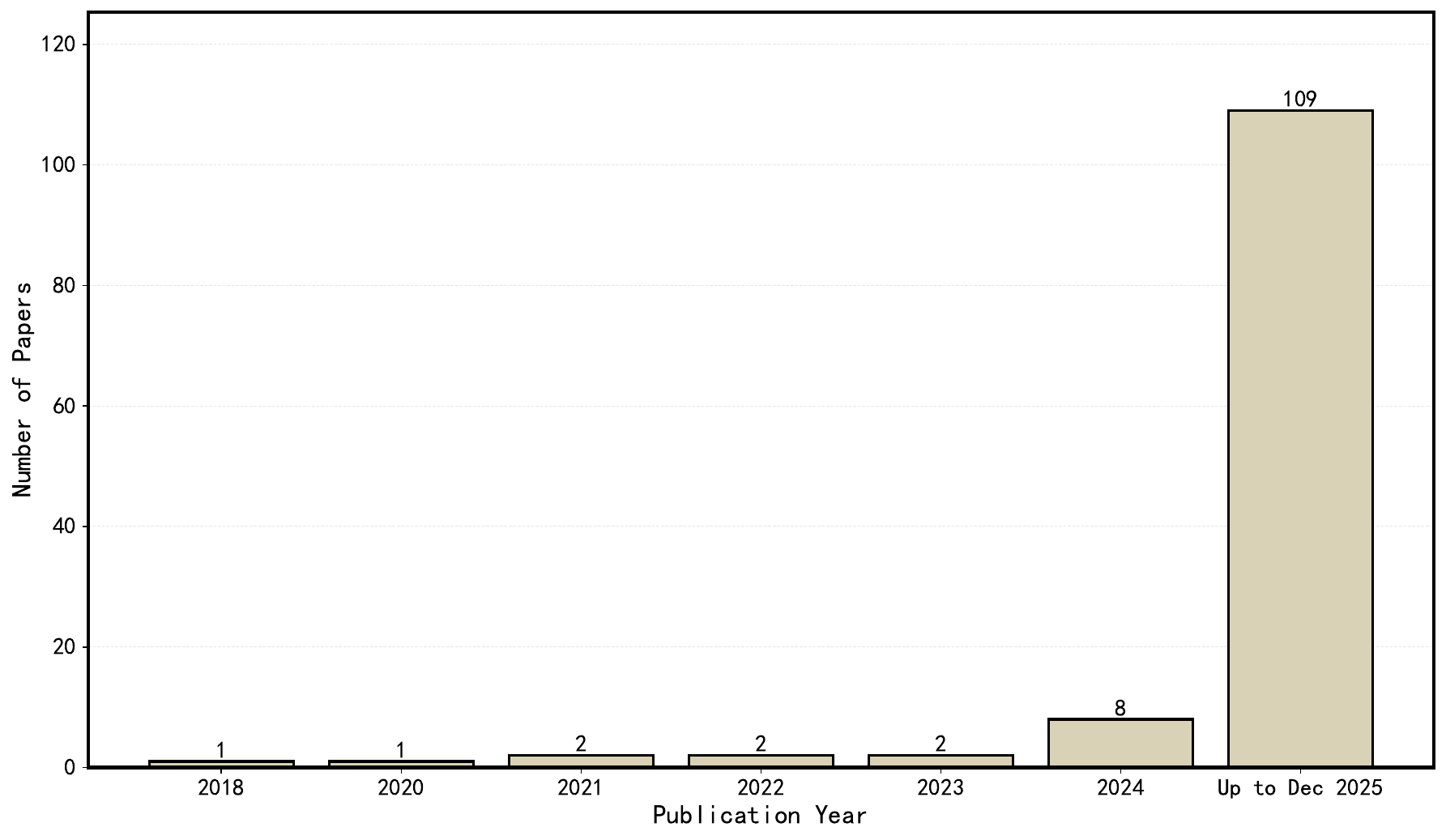}
    \caption{Distribution on publication year of surveyed papers.}
    \label{fig:year_distribution}
\end{figure}

To show the research momentum in data-efficient reinforcement learning for large reasoning models (LLMs), we conduct a statistical analysis of the publication year of all surveyed papers. As shown in Figure~\ref{fig:year_distribution}, the field has experienced an extraordinary growth trajectory. Early works are scarce, with $1$ publication in $2018$ and $2020$ each, followed by $2$ papers in both $2021$ and $2022$, and $2$ papers in $2023$. The field witnessed a modest increase to $8$ papers in $2024$, before experiencing an explosive surge to $109$ publications by December $2025$. This remarkable acceleration highlights the field's transition from an emerging research direction to a mainstream paradigm in LLM development.

\begin{figure}[t]
    \centering
    \includegraphics[width=\linewidth]{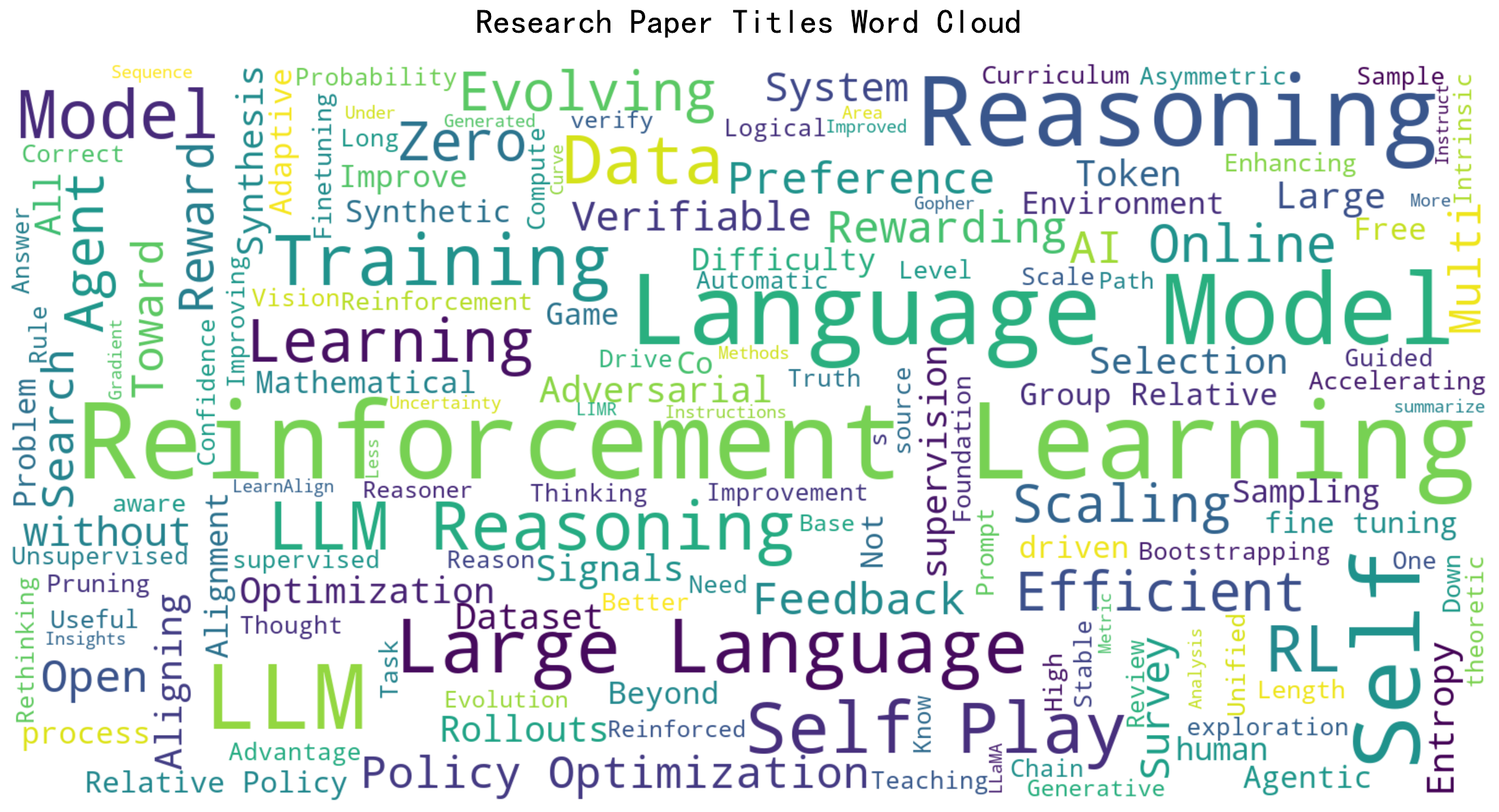}
    \caption{Word cloud of research paper titles.}
    \label{fig:wordcloud}
\end{figure}

Furthermore, we perform a word frequency analysis on the titles of all surveyed papers. Figure~\ref{fig:wordcloud} reveals the core methodological and conceptual themes in current research. Prominently, \textit{Reasoning} and \textit{Learning} emerge as the most central terms, which reflect the field's primary focus on enhancing reasoning capabilities through learning mechanisms. The significant presence of \textit{Reinforcement}, \textit{Training}, \textit{Data} and \textit{Model} underscores the importance of reinforcement learning based training strategies. Additionally, key concepts such as \textit{Self}, \textit{Evolving}, and \textit{Agent} highlight the growing emphasis on self-improvement and agentic frameworks, and terms such as \textit{Verifiable} and \textit{Rewarding} also indicate the field's focus on reward engineering. This visualization demonstrates the rapid advances that data-efficient RL for large reasoning models has seen in recent years, laying a foundation for future progress in LLM capabilities.

\section{Acknowledgment of AI Assistance in Writing and Revision}
\label{app:acknowledgement_AI}
We only use large language models for grammar checking and language refinement, and we strictly adhere to the ACL policy on AI writing assistance. All ideas and technical content in this paper are original contributions of the authors.

\section{Further Analysis}

\subsection{Discussion on Data Pruning}
\noindent \textbf{\textit{Uncertainty-Aware Pruning.}} Uncertainty-aware pruning uses epistemic uncertainty to identify samples where the model’s knowledge is most insufficient, concentrating the training budget on informative data. UFO-RL~\cite{zhao2025uforluncertaintyfocusedoptimizationefficient} computes a single-pass confidence score from average token-level log-probabilities and selects samples within the model’s zone of proximal development, matching full-data performance with only 10\% of data. ReST~\cite{gulcehre2023reinforcedselftrainingrestlanguage} iteratively grows the training set by generating candidates from the current policy and filtering them through a reward threshold, coupling data curation with policy improvement. BAL-PM~\cite{NEURIPS2024_d5e256c9} combines ensemble-based epistemic uncertainty of the reward model with a prompt-distribution entropy term for active preference data acquisition, reducing required labels by 33–68\%. ADPO~\cite{Ji2024ReinforcementLF} formulates RLHF as a contextual dueling bandit and introduces an uncertainty-aware query criterion that requests human feedback only when the reward gap is poorly estimated, halving query complexity while matching DPO performance. Uncertainty-Penalized DPO~\cite{houliston2024uncertaintypenalizeddirectpreferenceoptimization} attenuates the DPO loss gradient for preference pairs with high reward uncertainty, down-weighting ambiguous or mislabeled samples.

\subsection{Discussion on Reward Engineering}
\noindent \textbf{\textit{Theoretical Perspectives on Internal Feedback.}} Recent work has begun to provide theoretical grounding for why internal feedback signals can substitute for external rewards.~\citet{zhang2025no} formally investigate Reinforcement Learning from Internal Feedback, showing that token-level entropy minimization, trajectory-level entropy minimization, and self-certainty maximization are partially equivalent optimization objectives under mild assumptions. CoVo~\cite{zhang2025consistentpathsleadtruth} provides a complementary theoretical perspective by reinterpreting its consistency-volatility reward as a variational inference objective, treating reasoning trajectories as latent variables. This formulation grounds the self-rewarding mechanism in both reasoning paths and final answers, offering a principled explanation for why correct responses exhibit convergent trajectory patterns while incorrect ones diverge.

From a broader perspective,~\citet{shao2025spurious} present a striking empirical finding with theoretical implications: RLVR can elicit strong mathematical reasoning even with spurious rewards on certain model families.~\citet{chen2025reshapingreasoningllmstheoretical} further support this point by a two-stage mathematical framework modeling reasoning as $q \to r \to a$ (question-reason-answer). For RLVR, they prove convergence to the reasoning pattern with the highest success rate. For RLIF, their analysis explains both the initial performance gains and eventual degradation observed empirically.~\citet{agarwal2025unreasonable} further demonstrate the unreasonable effectiveness of entropy minimization: using negative entropy as the sole RL reward, their EM-RL achieves performance comparable to supervised GRPO and RLOO on challenging reasoning benchmarks. These theoretical and empirical results collectively suggest a key insight that under data scarcity, RL may function less as a mechanism for acquiring new capabilities and more as a redistribution mechanism that sharpens the policy over reasoning patterns already learned during pretraining.

\noindent \textbf{\textit{Adaptive Multi-Objective Trade-offs.}} Under data scarcity, static reward weighting fails without sufficient validation data for tuning. Effective reward engineering thus requires denser signals and adaptive mechanisms to balance conflicting objectives. FINE-GRAINED RLHF~\cite{NEURIPS2023_b8c90b65} replaces holistic feedback with segment-level, multi-type rewards, yielding denser supervision and higher sample efficiency. Safe-RLHF~\cite{dai2023saferlhfsafereinforcement} decouples helpfulness and harmlessness into reward/cost models, using Lagrangian constraints for dynamic balancing without hyperparameter searches. MAESTRO~\cite{zhao2026maestrometalearningadaptiveestimation} treats scalarization as a dynamic latent policy, co-evolving weights via group-relative meta-signals to eliminate manual tuning. Extending this paradigm, PAMA~\cite{10.1007/978-3-032-06078-5_15} transforms multi-objective RLHF into a convex optimization with closed-form Pareto convergence at $O(n)$ complexity. JobRec~\cite{kan2026deconflatingpreferencequalificationconstrained} applies similar Lagrangian constraints to de-conflate candidate preference and employer qualification under data scarcity. IB-GRPO~\cite{wang2026ibgrpoaligningllmbasedlearning} employs $\epsilon$-dominance indicators to compute group-relative advantages across multiple objectives, avoiding manual scalarization while leveraging hybrid expert demonstrations. Collectively, these methods shift reward engineering from static design toward adaptive, data-efficient mechanisms for navigating conflicting objectives without extensive validation data.

\section{Literature Review Summary}
\label{app:lit_review}
To provide an overview of the literature examined in this study, we present a detailed summary table in this appendix covering all discussed works in our analysis. Each entry includes six columns : \textbf{Title} (the complete publication name); \textbf{Section} and \textbf{Subsection} (place in the three-level taxonomy); \textbf{Year} (the publication date); \textbf{Venue} (the publication outlet) and \textbf{Link} (URL to the original source of paper), as shown in Table~\ref{tab:papers}.

\setlength{\tabcolsep}{3pt} 
\renewcommand{\arraystretch}{1.7} 

\newcolumntype{C}[1]{>{\centering\arraybackslash}m{#1}}

\clearpage
{
\scriptsize
\onecolumn
\begin{longtable}{C{5cm}C{2cm}C{2cm}C{1cm}C{2cm}C{0.8cm}}
\caption{\normalsize Summary of Referenced Papers}
\label{tab:papers} \\
\hline
\textbf{Title} & \textbf{Section} & \textbf{Subsection} & \textbf{Year} & \textbf{Venue} & \textbf{Link} \\
\hline
\endfirsthead

\multicolumn{6}{c}%
{\normalsize \tablename\ \thetable\ -- Continued} \\
\hline
\textbf{Title} & \textbf{Section} & \textbf{Subsection} & \textbf{Year} & \textbf{Venue} & \textbf{Link} \\
\hline
\endhead

\hline
\multicolumn{6}{r}{\textit{ \normalsize Continued on next page}} \\
\endfoot

\hline
\endlastfoot

Uncertainty Under the Curve: A Sequence-Level Entropy Area Metric for Reasoning LLM & Level 1: Data-Centric Perspective & Data Pruning & 2025 & Arxiv & \href{https://arxiv.org/abs/2508.20384}{link} \\
\hline
LLaMA: Open and Efficient Foundation Language Models & Level 1: Data-Centric Perspective & Data Pruning & 2025 & Arxiv & \href{https://arxiv.org/abs/2302.13971}{link} \\
\hline
Learning to summarize from human feedback & Level 1: Data-Centric Perspective & Data Pruning & 2020 & NeurIPS & \href{https://dl.acm.org/doi/abs/10.5555/3495724.3495977}{link} \\
\hline
Scaling Language Models: Methods, Analysis \& Insights from Training Gopher & Level 1: Data-Centric Perspective & Data Pruning & 2021 & Arxiv & \href{https://arxiv.org/abs/2112.11446}{link} \\
\hline
Self-Instruct: Aligning Language Models with Self-Generated Instructions & Level 1: Data-Centric Perspective & Data Pruning & 2023 & ACL & \href{https://aclanthology.org/2023.acl-long.754/}{link} \\
\hline
LIMR: Less is More for RL Scaling & Level 1: Data-Centric Perspective & Data Pruning & 2025 & Arxiv & \href{https://arxiv.org/abs/2502.11886}{link} \\
\hline
LearnAlign: Reasoning Data Selection for Reinforcement Learning in Large Language Models Based on Improved Gradient Alignment & Level 1: Data-Centric Perspective & Data Pruning & 2025 & Arxiv & \href{https://arxiv.org/abs/2506.11480}{link} \\
\hline
Unified Data Selection for LLM Reasoning & Level 1: Data-Centric Perspective & Data Pruning & 2025 & ICLR & \href{https://openreview.net/forum?id=heVn5cNfje}{link} \\
\hline
LSPO: Length-aware Dynamic Sampling for Policy Optimization in LLM Reasoning & Level 1: Data-Centric Perspective & Data Pruning & 2025 & Arxiv & \href{https://arxiv.org/abs/2510.01459}{link} \\
\hline
Can Prompt Difficulty be Online Predicted for Accelerating RL Finetuning of Reasoning Models? & Level 1: Data-Centric Perspective & Data Pruning & 2025 & KDD & \href{https://arxiv.org/abs/2507.04632}{link} \\
\hline
BOTS: A Unified Framework for Bayesian Online Task Selection in LLM Reinforcement Finetuning & Level 1: Data-Centric Perspective & Data Pruning & 2025 & Arxiv & \href{https://arxiv.org/abs/2510.26374}{link} \\
\hline
Angles Don't Lie: Unlocking Training?Efficient RL Through the Model's Own Signals & Level 1: Data-Centric Perspective & Data Pruning & 2025 & NeurIPS & \href{https://arxiv.org/abs/2506.02281}{link} \\
\hline
Online Difficulty Filtering for Reasoning Oriented Reinforcement Learning & Level 1: Data-Centric Perspective & Data Pruning & 2025 & Arxiv & \href{https://arxiv.org/abs/2504.03380v1}{link} \\
\hline
SPEED-RL: Faster Training of Reasoning Models via Online Curriculum Learning & Level 1: Data-Centric Perspective & Data Pruning & 2025 & ICML Workshop & \href{https://arxiv.org/abs/2506.09016}{link} \\
\hline
Secrets of RLHF in Large Language Models Part I: PPO &  Level 1: Data-Centric Perspective & Data Pruning & 2023 & NeurIPS Workshop & \href{https://arxiv.org/abs/2307.04964}{link} \\
\hline
Influence Functions for Preference Dataset Pruning & Level 1: Data-Centric Perspective & Data Pruning & 2025 & NeurIPS Workshop & \href{https://arxiv.org/abs/2507.14344}{link} \\
\hline
CPPO: Accelerating the Training of Group Relative Policy Optimization-Based Reasoning Models & Level 1: Data-Centric Perspective & Data Pruning & 2025 & NeurIPS & \href{https://arxiv.org/abs/2503.22342v1}{link} \\
\hline
Not All Rollouts are Useful: Down-Sampling Rollouts in LLM Reinforcement Learning & Level 1: Data-Centric Perspective & Data Pruning & 2025 & Arxiv & \href{https://arxiv.org/abs/2504.13818}{link} \\
\hline
Enhancing Chat Language Models by Scaling High-quality Instructional Conversations & Level 1: Data-Centric Perspective & Data Synthesis & 2023 & EMNLP & \href{https://aclanthology.org/2023.emnlp-main.183/}{link} \\
\hline
STaR: Bootstrapping Reasoning With Reasoning & Level 1: Data-Centric Perspective & Data Synthesis & 2022 & NeurIPS & \href{https://papers.nips.cc/paper_files/paper/2022/hash/639a9a172c044fbb64175b5fad42e9a5-Abstract-Conference.html}{link} \\
\hline
Constitutional AI: Harmlessness from AI Feedback & Level 1: Data-Centric Perspective & Data Synthesis & 2022 & Arxiv & \href{https://arxiv.org/abs/2212.08073}{link} \\
\hline
ULTRAFEEDBACK: Boosting Language Models with Scaled AI Feedback & Level 1: Data-Centric Perspective & Data Synthesis & 2024 & ICML & \href{https://proceedings.mlr.press/v235/cui24f.html}{link} \\
\hline
CodeUltraFeedback: An LLM-as-a-Judge Dataset for Aligning Large Language Models to Coding Preferences & Level 1: Data-Centric Perspective & Data Synthesis & 2025 & ACM Transactions on Software Engineering and Methodology & \href{https://dl.acm.org/doi/10.1145/3736407}{link} \\
\hline
Fair-PP: A Synthetic Dataset for Aligning LLM with Personalized Preferences of Social Equity & Level 1: Data-Centric Perspective & Data Synthesis & 2025 & Arxiv & \href{https://arxiv.org/abs/2505.11861}{link} \\
\hline
The Fellowship of the LLMs: Multi-Model Workflows for Synthetic Preference Optimization Dataset Generation & Level 1: Data-Centric Perspective & Data Synthesis & 2025 & GEM & \href{https://aclanthology.org/2025.gem-1.4/}{link} \\
\hline
Reasoning Gym: Reasoning Environments for Reinforcement Learning with Verifiable Rewards & Level 1: Data-Centric Perspective & Data Synthesis & 2025 & NeurIPS & \href{https://arxiv.org/abs/2505.24760}{link} \\
\hline
Enigmata: Scaling Logical Reasoning in Large Language Models with Synthetic Verifiable Puzzles & Level 1: Data-Centric Perspective & Data Synthesis & 2025 & Arxiv & \href{https://arxiv.org/abs/2505.19914}{link} \\
\hline
SynLogic: Synthesizing Verifiable Reasoning Data at Scale for Learning Logical Reasoning and Beyond & Level 1: Data-Centric Perspective & Data Synthesis & 2025 & NeurIPS & \href{https://huggingface.co/papers/2505.19641}{link} \\
\hline
EvoSyn: Generalizable Evolutionary Data Synthesis for Verifiable Learning & Level 1: Data-Centric Perspective & Data Synthesis & 2025 & Arxiv & \href{https://arxiv.org/abs/2510.17928}{link} \\
\hline
SynthRL: Scaling Visual Reasoning with Verifiable Data Synthesis & Level 1: Data-Centric Perspective & Data Synthesis & 2025 & Arxiv & \href{https://arxiv.org/abs/2506.02096}{link} \\
\hline
Making Mathematical Reasoning Adaptive & Level 1: Data-Centric Perspective & Data Synthesis & 2025 & Arxiv & \href{https://arxiv.org/abs/2510.04617}{link} \\
\hline
SwS: Self-aware Weakness-driven Problem Synthesis in Reinforcement Learning for LLM Reasoning & Level 1: Data-Centric Perspective & Data Synthesis & 2025 & NeurIPS & \href{https://arxiv.org/abs/2506.08989}{link} \\
\hline
Reward-Guided Prompt Evolving in Reinforcement Learning for LLMs & Level 1: Data-Centric Perspective & Data Synthesis & 2025 & ICML & \href{https://proceedings.mlr.press/v267/ye25a.html}{link} \\
\hline
Direct Advantage Regression: Aligning LLMs with Online AI Reward & Level 1: Data-Centric Perspective & Data Synthesis & 2025 & Arxiv & \href{https://arxiv.org/abs/2504.14177}{link} \\
\hline
Online Self-Preferring Language Models & Level 1: Data-Centric Perspective & Data Synthesis & 2024 & Arxiv & \href{https://arxiv.org/abs/2405.14103}{link} \\
\hline
LoopTool: Closing the Data–Training Loop for Robust LLM Tool Calls & Level 1: Data-Centric Perspective & Data Synthesis & 2025 & Arxiv & \href{https://arxiv.org/abs/2511.09148}{link} \\
\hline
WebRL: Training LLM Web Agents via Self-Evolving Online Curriculum Reinforcement Learning & Level 1: Data-Centric Perspective & Data Synthesis & 2025 & ICLR & \href{https://arxiv.org/abs/2411.02337}{link} \\
\hline
Shallow Preference Signals: Large Language Model Aligns Even Better with Truncated Data? & Level 1: Data-Centric Perspective & Data Compression & 2025 & GEM & \href{https://aclanthology.org/2025.gem-1.48/}{link} \\
\hline
Beyond the 80/20 Rule: High-Entropy Minority Tokens Drive Effective Reinforcement Learning for LLM Reasoning & Level 1: Data-Centric Perspective & Data Compression & 2025 & NeurIPS & \href{https://openreview.net/forum?id=yfcpdY4gMP}{link} \\
\hline
Dapo: An open-source llm reinforcement learning system at scale & Level 1: Data-Centric Perspective & Data Compression & 2025 & NeurIPS & \href{https://openreview.net/forum?id=2a36EMSSTp}{link} \\
\hline
Beyond Correctness: Harmonizing Process and Outcome Rewards through RL Training & Level 1: Data-Centric Perspective & Data Compression & 2025 & NeurIPS & \href{https://openreview.net/forum?id=i6lySxAqe4}{link} \\
\hline
Reinforcement Learning for Reasoning in Large Language Models with One Training Example & Level 1: Data-Centric Perspective & Data Compression & 2025 & NeurIPS & \href{https://arxiv.org/abs/2504.20571}{link} \\
\hline
Sharpness-Controlled Group Relative Policy Optimization with Token-Level Probability Shaping & Level 1: Data-Centric Perspective & Data compression & 2025 & Arxiv & \href{https://arxiv.org/abs/2511.00066}{link} \\
\hline
One Sample to Rule Them All: Extreme Data Efficiency in RL Scaling & Level 1: Data-Centric Perspective & Data Compression & 2025 & Arxiv & \href{https://openreview.net/forum?id=Hc1HsOJu4k}{link} \\
\hline
Do Not Let Low-Probability Tokens Over-Dominate in RL for LLMs & Level 1: Data-Centric Perspective & Data Compression & 2025 & NeurIPS & \href{https://openreview.net/forum?id=FeCCRol2zo}{link} \\
\hline
Token-Regulated Group Relative Policy Optimization for Stable Reinforcement Learning in Large Language Models & Level 1: Data-Centric Perspective & Data Compression & 2025 & Arxiv & \href{https://openreview.net/forum?id=rhV6QTMqq1}{link} \\
\hline
Towards Flash Thinking via Decoupled Advantage Policy Optimization & Level 1: Data-Centric Perspective & Data Compression & 2025 & Arxiv & \href{https://arxiv.org/abs/2510.15374}{link} \\
\hline
DAST: Difficulty-Adaptive Slow-Thinking for Large Reasoning Models & Level 1: Data-Centric Perspective & Data Compression & 2025 & EMNLP & \href{https://arxiv.org/abs/2503.04472}{link} \\
\hline
ThinkPrune: Pruning Long Chain-of-Thought of LLMs via Reinforcement Learning & Level 1: Data-Centric Perspective & Data Compression & 2025 & TMLR & \href{https://openreview.net/forum?id=V51gPu1uQD}{link} \\
\hline
S-GRPO: Early Exit via Reinforcement Learning in Reasoning Models & Level 1: Data-Centric Perspective & Data Compression & 2025 & NeurIPS & \href{https://openreview.net/forum?id=wNMK5o0Vfg}{link} \\
\hline
Interleaved Reasoning for Large Language Models via Reinforcement Learning & Level 1: Data-Centric Perspective & Data Compression & 2025 & Arxiv & \href{https://openreview.net/forum?id=DIWdk9Zo7g}{link} \\
\hline
Compressing Chain-of-Thought in LLMs via Step Entropy & Level 1: Data-Centric Perspective & Data Compression & 2025 & Arxiv & \href{https://arxiv.org/abs/2508.03346}{link} \\
\hline
Sample-efficient LLM Optimization with Reset Replay & Level 2: Training-Centric Perspective & Policy Optimization & 2025 & Arxiv & \href{https://arxiv.org/pdf/2508.06412}{link} \\
\hline
Inpainting-Guided Policy Optimization for Diffusion Large Language Models & Level 2: Training-Centric Perspective & Policy Optimization & 2025 & Arxiv & \href{https://arxiv.org/pdf/2509.10396?}{link} \\
\hline
KTO: Model alignment as prospect theoretic optimization & Level 2: Training-Centric Perspective & Policy Optimization & 2024 & ICML & \href{https://arxiv.org/pdf/2402.01306}{link} \\
\hline
Can Large Reasoning Models Self-Train? & Level 2: Training-Centric Perspective & Reward Engineering & 2025 & Arxiv & \href{https://arxiv.org/pdf/2505.21444?}{link} \\
\hline
Test-Time Reinforcement Learning & Level 2: Training-Centric Perspective & Reward Engineering & 2025 & Arxiv & \href{https://arxiv.org/pdf/2504.16084}{link} \\
\hline
Semantic Voting: A Self-Evaluation-Free Approach for Efficient LLM Self-Improvement on Unverifiable Open-ended Tasks & Level 2: Training-Centric Perspective & Reward Engineering & 2025 & NeurIPS & \href{https://arxiv.org/pdf/2509.23067}{link} \\
\hline
Wisdom of the Crowd: Reinforcement Learning from Coevolutionary Collective Feedback & Level 2: Training-Centric Perspective & Reward Engineering & 2025 & Arxiv & \href{https://arxiv.org/pdf/2508.12338}{link} \\
\hline
First SFT, Second RL, Third UPT: Continual Improving Multi-Modal LLM Reasoning via Unsupervised Post-Training & Level 2: Training-Centric Perspective & Reward Engineering & 2025 & NeurIPS & \href{https://openreview.net/pdf?id=HL1j92hb6z}{link} \\
\hline
Learning to Reason without External Rewards & Level 2: Training-Centric Perspective & Reward Engineering & 2025 & Arxiv & \href{https://arxiv.org/abs/2505.19590}{link} \\
\hline
KnowRL: Teaching Language Models to Know What They Know & Level 2: Training-Centric Perspective & Reward Engineering & 2025 & Arxiv & \href{https://arxiv.org/pdf/2510.11407?}{link} \\
\hline
Self Rewarding Self Improving & Level 2: Training-Centric Perspective & Reward Engineering & 2025 & Arxiv & \href{https://arxiv.org/pdf/2505.08827}{link} \\
\hline
Co-rewarding: Stable Self-supervised RL for Eliciting Reasoning in Large Language Models & Level 2: Training-Centric Perspective & Reward Engineering & 2025 & Arxiv & \href{https://arxiv.org/pdf/2508.00410}{link} \\
\hline
Spurious rewards: Rethinking training signals in rlvr & Level 2: Training-Centric Perspective & Reward Engineering & 2025 & Arxiv & \href{https://arxiv.org/pdf/2506.10947?}{link} \\
\hline
Surrogate Signals from Format and Length: Reinforcement Learning for Solving Mathematical Problems without Ground Truth Answers & Level 2: Training-Centric Perspective & Reward Engineering & 2025 & Arxiv & \href{https://arxiv.org/pdf/2505.19439}{link} \\
\hline
No Free Lunch: Rethinking Internal Feedback for LLM Reasoning & Level 2: Training-Centric Perspective & Reward Engineering & 2025 & Arxiv & \href{https://arxiv.org/pdf/2506.17219}{link} \\
\hline
CoMAS: Co-Evolving Multi-Agent Systems via Interaction Rewards & Level 2: Training-Centric Perspective & Reward Engineering & 2025 & Arxiv & \href{https://arxiv.org/pdf/2510.08529?}{link} \\
\hline
Compute as teacher: Turning inference compute into reference-free supervision & Level 2: Training-Centric Perspective & Reward Engineering & 2025 & Arxiv & \href{https://arxiv.org/pdf/2509.14234}{link} \\
\hline
Self-Play Preference Optimization for Language Model Alignment & Level 2: Training-Centric Perspective & Reward Engineering & 2025 & ICLR & \href{https://openreview.net/pdf?id=a3PmRgAB5T}{link} \\
\hline
Right Question is Already Half the Answer: Fully Unsupervised LLM Reasoning Incentivization & Level 2: Training-Centric Perspective & Reward Engineering & 2025 & NeurIPS & \href{https://arxiv.org/abs/2504.05812}{link} \\
\hline
Consistent Paths Lead to Truth: Self-Rewarding Reinforcement Learning for LLM Reasoning & Level 2: Training-Centric Perspective & Reward Engineering & 2025 & Arxiv & \href{https://arxiv.org/abs/2506.08745}{link} \\
\hline
Evolving language models without labels: Majority drives selection, novelty promotes variation & Level 2: Training-Centric Perspective & Reward Engineering & 2025 & Arxiv & \href{https://arxiv.org/abs/2509.15194}{link} \\
\hline
Learning to think: Information-theoretic reinforcement fine-tuning for llms & Level 2: Training-Centric Perspective & Reward Engineering & 2025 & Arxiv & \href{https://arxiv.org/abs/2505.10425}{link} \\
\hline
Semi-supervised Fine-tuning for Large Language Models
 & Level 2: Training-Centric Perspective & Reward Engineering & 2025 & *ACL & \href{https://arxiv.org/abs/2410.14745}{link} \\
\hline
Maximizing Confidence Alone Improves Reasoning & Level 2: Training-Centric Perspective & Reward Engineering & 2025 & Arxiv & \href{https://arxiv.org/abs/2505.22660}{link} \\
\hline
Confidence Is All You Need: Few-Shot RL Fine-Tuning of Language Models & Level 2: Training-Centric Perspective & Reward Engineering & 2025 & Arxiv & \href{https://arxiv.org/abs/2506.06395}{link} \\
\hline
The unreasonable effectiveness of entropy minimization in llm reasoning & Level 2: Training-Centric Perspective & Reward Engineering & 2025 & Arxiv & \href{https://arxiv.org/abs/2505.15134}{link} \\
\hline
Cde: Curiosity-driven exploration for efficient reinforcement learning in large language models & Level 2: Training-Centric Perspective & Reward Engineering & 2025 & Arxiv & \href{https://arxiv.org/abs/2509.09675}{link} \\
\hline
Training Language Models to Self-Correct via Reinforcement Learning & Level 2: Training-Centric Perspective & Reward Engineering & 2025 & ICLR & \href{https://arxiv.org/abs/2409.12917}{link} \\
\hline
ReviewRL: Towards Automated Scientific Review with RL & Level 2: Training-Centric Perspective & Reward Engineering & 2025 & *ACL & \href{https://arxiv.org/abs/2508.10308}{link} \\
\hline
AI-powered peer review needs human supervision & Level 2: Training-Centric Perspective & Reward Engineering & 2025 & Journal of Information, Communication and Ethics in Society & \href{https://www.emerald.com/jices/article-abstract/23/1/104/1244930/AI-powered-peer-review-needs-human-supervision?redirectedFrom=fulltext}{link} \\
\hline
AlphaZero-Like Tree-Search can Guide Large Language Model Decoding and Training & Level 2: Training-Centric Perspective & Trajectory Generation & 2024 & ICML & \href{https://openreview.net/pdf?id=fLO9VaAb3B}{link} \\
\hline
Alphamath almost zero: process supervision without process & Level 2: Training-Centric Perspective & Trajectory Generation & 2024 & NeurIPS & \href{https://openreview.net/pdf?id=VaXnxQ3UKo}{link} \\
\hline
Toward Efficient Exploration by Large Language Model Agents & Level 2: Training-Centric Perspective & Trajectory Generation & 2025 & ICML Workshop & \href{https://arxiv.org/abs/2504.20997}{link} \\
\hline
Search-R1: Training LLMs to Reason and Leverage Search Engines with Reinforcement Learning
 & Level 2: Training-Centric Perspective & Trajectory Generation & 2025 & Arxiv & \href{https://arxiv.org/abs/2503.09516}{link} \\
\hline
Not all rollouts are useful: Down-sampling rollouts in llm reinforcement learning & Level 2: Training-Centric Perspective & Trajectory Generation & 2025 & Arxiv & \href{https://arxiv.org/abs/2504.13818}{link} \\
\hline
Intrinsic Motivation and Automatic Curricula via Asymmetric Self-Play & Level 3: Framework-Centric Perspective & Asymmetric Co-Evolution & 2018 & ICLR & \href{https://openreview.net/forum?id=SkT5Yg-RZ}{link} \\
\hline
Asymmetric self-play for automatic goal discovery in robotic manipulation & Level 3: Framework-Centric Perspective & Asymmetric Co-Evolution & 2021 & Arxiv & \href{https://arxiv.org/abs/2101.04882}{link} \\
\hline
Self-Questioning Language Models & Level 3: Framework-Centric Perspective & Asymmetric Co-Evolution & 2025 & Arxiv & \href{https://arxiv.org/abs/2508.03682}{link} \\
\hline
Learning to Pose Problems: Reasoning-Driven and Solver-Adaptive Data Synthesis for Large Reasoning Models & Level 3: Framework-Centric Perspective & Asymmetric Co-Evolution & 2025 & Arxiv & \href{https://arxiv.org/abs/2511.09907}{link} \\
\hline
Better LLM Reasoning via Dual-Play & Level 3: Framework-Centric Perspective & Asymmetric Co-Evolution & 2025 & Arxiv & \href{https://arxiv.org/abs/2511.11881}{link} \\
\hline
VisPlay: Self-Evolving Vision-Language Models from Images & Level 3: Framework-Centric Perspective & Asymmetric Co-Evolution & 2025 & Arxiv & \href{https://arxiv.org/abs/2511.15661}{link} \\
\hline
Search Self-play: Pushing the Frontier of Agent Capability without Supervision & Level 3: Framework-Centric Perspective & Asymmetric Co-Evolution & 2025 & Arxiv & \href{https://arxiv.org/abs/2510.18821}{link} \\
\hline
SPICE: Self-Play In Corpus Environments Improves Reasoning & Level 3: Framework-Centric Perspective & Asymmetric Co-Evolution & 2025 & Arxiv & \href{https://arxiv.org/abs/2510.24684}{link} \\
\hline
AceSearcher: Bootstrapping Reasoning and Search for LLMs via Reinforced Self-Play
 & Level 3: Framework-Centric Perspective & Asymmetric Co-Evolution & 2025 & NeurIPS & \href{https://openreview.net/forum?id=jSgCM0uZn3}{link} \\
\hline
ZeroGUI: Automating Online GUI Learning at Zero Human Cost & Level 3: Framework-Centric Perspective & Self-evolving
Framework & 2025 & Arxiv & \href{https://arxiv.org/abs/2505.23762}{link} \\
\hline
The Path of Self-Evolving Large Language Models: Achieving Data-Efficient Learning via Intrinsic Feedback & Level 3: Framework-Centric Perspective & Self-evolving
Framework & 2025 & Arxiv & \href{https://arxiv.org/abs/2510.02752}{link} \\
\hline
Towards Agentic Self-Learning LLMs in Search Environment & Level 3: Framework-Centric Perspective & Self-evolving
Framework & 2025 & Arxiv & \href{https://arxiv.org/abs/2510.14253}{link} \\
\hline
Absolute Zero: Reinforced Self-play Reasoning with Zero Data & Level 3: Framework-Centric Perspective & Self-evolving
Framework & 2025 & Arxiv & \href{https://arxiv.org/abs/2505.03335}{link} \\
\hline
An Extremely Data-efficient and Generative LLM-based Reinforcement Learning Agent for Recommenders & Level 3: Framework-Centric Perspective & Self-evolving
Framework & 2024 & Arxiv & \href{https://arxiv.org/abs/2408.16032}{link} \\
\hline
Self-rewarding correction for mathematical reasoning & Level 3: Framework-Centric Perspective & Self-evolving
Framework & 2025 & Arxiv & \href{https://arxiv.org/abs/2502.19613}{link} \\
\hline
S2R: Teaching LLMs to Self-verify and Self-correct via Reinforcement Learning & Level 3: Framework-Centric Perspective & Self-evolving
Framework & 2025 & Arxiv & \href{https://arxiv.org/abs/2502.12853}{link} \\
\hline
SSR-Zero: Simple Self-Rewarding Reinforcement Learning for Machine Translation & Level 3: Framework-Centric Perspective & Self-evolving
Framework & 2025 & Arxiv & \href{https://arxiv.org/abs/2505.16637}{link} \\
\hline
Diagnosing and Mitigating System Bias in Self-Rewarding RL & Level 3: Framework-Centric Perspective & Self-evolving
Framework & 2025 & Arxiv & \href{https://arxiv.org/abs/2510.08977}{link} \\
\hline
SeRL: Self-Play Reinforcement Learning for Large Language Models with Limited Data & Level 3: Framework-Centric Perspective & Self-evolving
Framework & 2025 & Arxiv & \href{https://arxiv.org/abs/2505.20347}{link} \\
\hline

SPC: Evolving Self-Play Critic via Adversarial Games for LLM Reasoning & Level 3: Framework-Centric Perspective & Asymmetric Co-Evolution Frameworks & 2025 & NeurIPS & \href{https://arxiv.org/abs/2504.19162}{link} \\
\hline
Generative Adversarial Reasoner: Enhancing LLM Reasoning with Adversarial Reinforcement Learning & Level 3: Framework-Centric Perspective & Asymmetric Co-Evolution Frameworks & 2025 & Arxiv & \href{https://arxiv.org/abs/2512.16917}{link} \\
\hline
Toward Training Superintelligent Software Agents through Self-Play SWE-RL & Level 3: Framework-Centric Perspective & Asymmetric Co-Evolution Frameworks & 2025 & Arxiv & \href{https://arxiv.org/abs/2512.18552}{link} \\
\hline
Self-playing Adversarial Language Game Enhances LLM Reasoning & Level 3: Framework-Centric Perspective & Asymmetric Co-Evolution Frameworks & 2025 & Arxiv & \href{https://arxiv.org/abs/2404.10642}{link} \\
\hline
SPIRAL: Self-Play on Zero-Sum Games Incentivizes Reasoning via Multi-Agent Multi-Turn Reinforcement Learning & Level 3: Framework-Centric Perspective & Multi-Agent Evolution Frameworks & 2025 & Arxiv & \href{https://arxiv.org/abs/2506.24119}{link} \\
\hline
Vision-Zero: Scalable VLM Self-Improvement via Strategic Gamified Self-Play & Level 3: Framework-Centric Perspective & Multi-Agent Evolution Frameworks & 2025 & Arxiv & \href{https://arxiv.org/abs/2509.25541}{link} \\
\hline
Propose, Solve, Verify: Self-Play Through Formal Verification & Level 3: Framework-Centric Perspective & Multi-Agent Evolution Frameworks & 2025 & Arxiv & \href{https://arxiv.org/abs/2512.18160}{link} \\
\hline
SPELL: Self-Play Reinforcement Learning for evolving Long-Context Language Models & Level 3: Framework-Centric Perspective & Multi-Agent Evolution Frameworks & 2025 & Arxiv & \href{https://arxiv.org/abs/2509.23863}{link} \\
\hline
Multi-Agent Evolve: LLM Self-Improve through Co-evolution & Level 3: Framework-Centric Perspective & Multi-Agent Evolution Frameworks & 2025 & Arxiv & \href{https://arxiv.org/abs/2510.23595}{link} \\
\hline
Meta-Learning Reinforcement Learning for Crypto-Return Prediction
 & Level 3: Framework-Centric Perspective & Multi-Agent Evolution Frameworks & 2025 & Arxiv & \href{https://arxiv.org/abs/2509.09751}{link} \\
\hline
Towards Understanding Self-play for LLM Reasoning & Level 3: Framework-Centric Perspective & Discussion & 2025 & Arxiv & \href{https://arxiv.org/abs/2510.27072}{link} \\
\hline

OpenAI o1 System Card & Introduction & Background & 2024 & Arxiv & \href{https://arxiv.org/abs/2412.16720}{link} \\
\hline
SimpleRL-Zoo: Investigating and Taming Zero Reinforcement Learning for Open Base Models in the Wild & Introduction & Background & 2025 & COLM & \href{https://openreview.net/forum?id=vSMCBUgrQj}{link} \\
\hline
Kimi k1.5: Scaling Reinforcement Learning with LLMs & Introduction & Background & 2025 & Arxiv & \href{https://arxiv.org/abs/2501.12599}{link} \\
\hline
Open-Reasoner-Zero: An Open Source Approach to Scaling Up Reinforcement Learning on the Base Model & Introduction & Background & 2025 & NeurIPS & \href{https://arxiv.org/abs/2503.24290}{link} \\
\hline
A survey of reinforcement learning for large reasoning models & Survey & Survey & 2025 & Arxiv & \href{https://arxiv.org/abs/2509.08827}{link} \\
\hline
The Landscape of Agentic Reinforcement Learning for LLMs: A Survey & Survey & Survey & 2025 & Arxiv & \href{https://arxiv.org/abs/2509.02547}{link} \\
\hline
A Survey on Efficient Large Language Model Training From Data-centric Perspectives & Survey & Survey & 2025 & ACL & \href{https://aclanthology.org/2025.acl-long.1493.pdf}{link} \\
\hline
A Survey on Self-Evolution of Large Language Models & Survey & Survey & 2024 & Arxiv & \href{https://arxiv.org/abs/2404.14387}{link} \\
\hline
A Comprehensive Survey of Self-Evolving AI Agents: A New Paradigm Bridging Foundation Models and Lifelong Agentic Systems & Survey & Survey & 2025 & Arxiv & \href{https://arxiv.org/abs/2508.07407}{link} \\
\hline
A Survey of Self-Evolving Agents: What, When, How, and Where to Evolve on the Path to Artificial Super Intelligence & Survey & Survey & 2026 & Arxiv & \href{https://arxiv.org/abs/2507.21046}{link} \\
\hline

\end{longtable}
}

\end{document}